\documentclass[12pt,a4paper]{article}

\usepackage{amsmath}
\usepackage{graphicx}
\usepackage{lipsum} 
\usepackage{geometry}
\usepackage{amssymb}
\usepackage{amsthm}

\usepackage{booktabs}
\usepackage{indentfirst}
\usepackage{anyfontsize}
\usepackage{verbatim}
\usepackage{newtxmath}

\usepackage{algorithm}
\usepackage{algpseudocode}

\usepackage{csquotes}

\usepackage{hyperref}
\usepackage{cleveref}
\hypersetup{colorlinks=true,linkcolor=blue,linktocpage,citecolor=black}

\usepackage[figuresright]{rotating}

\newcommand{\quat}{\mathbf{q}}
\newcommand{\Quat}{\mathbf{Q}}
\newcommand{\tr}{\mathrm{tr}}

\newcommand{\acos}{\mathrm{acos}\,}

\newcommand{\affilone}{Laboratoire de mathématiques Jean Leray, UMR CNRS 6629, Nantes Université, France.}
\newcommand{\affiltwo}{Laboratoire Lab-STICC, UMR CNRS 6285, Université Bretagne Sud, France.}
\newcommand{\affilthree}{CIC, Service de Neurologie, CR2TI-Inserm U1064, CHU de Nantes, France.}

\title{Generation of synthetic gait data: application to multiple sclerosis patients' gait pattern}
\author{
    \underline{Klervi Le Gall}\thanks{\affilone}
    \and 
    Lise Bellanger\thanks{\affiltwo}%
    \and
    David Laplaud\thanks{\affilthree}%
    \and
    Aymeric Stamm\footnotemark[1]
}

\date{\today}
\begin{document}

\theoremstyle{definition}
\newtheorem{definition}{Definition}[section]

\theoremstyle{plain}
\newtheorem{theorem}{Theorem}[section]

\theoremstyle{remark}
\newtheorem*{remark}{Remark}

\maketitle

\textbf{Corresponding author:} klervi.legall@univ-nantes.fr
\begin{abstract}
Multiple sclerosis (MS) is the leading cause of severe non-traumatic disability in young adults and its incidence is increasing worldwide. The variability of gait impairment in MS necessitates the development of a non-invasive, sensitive, and cost-effective tool for quantitative gait evaluation. The eGait movement sensor, designed to characterize human gait through unit quaternion time series (QTS) representing hip rotations, is a promising approach. However, the small sample sizes typical of clinical studies pose challenges for the stability of gait data analysis tools. To address these challenges, this article presents two key scientific contributions. First, a comprehensive framework is proposed for transforming QTS data into a form that preserves the essential geometric properties of gait while enabling the use of any tabular synthetic data generation method. Second, a synthetic data generation method is introduced, based on nearest neighbors weighting, which produces high-fidelity synthetic QTS data suitable for small datasets and private data environments. The effectiveness of the proposed method, is demonstrated through its application to MS gait data, showing very good fidelity and respect of the initial geometry of the data. Thanks to this work, we are able to produce synthetic data sets and work on the stability of clustering methods.
\end{abstract}


\section{Introduction} 

Multiple sclerosis (MS) is an autoimmune disease that affects the central nervous system by damaging the myelin sheath surrounding axons. This affects the transmission of electrical impulses leading to motor, sensory, cognitive, visual and sphincter disturbances~\cite{MS_symptoms}. MS is the leading cause of severe non-traumatic disability in young adults, with initial symptoms appearing around the age of $30$. Among the various symptoms, gait impairment is one of the most prevalent affecting $41\%$ of the MS population. It is considered by MS patients as the most problematic symptom as it marks the beginning of loss of autonomy which directly impacts quality of life~\cite{LaRocca_2011}.  

Gait impairment in MS is difficult to characterize due to its variability, making it essential to develop a non-invasive, sensitive, and cost-effective tool for quantitative gait evaluation in both MS patients and the general population. The eGait movement sensor, developed by the Jean Leray Mathematics Laboratory in Nantes (France), aims to characterize the human gait using unit quaternion time series (QTS) representing an average 3-dimensional hip rotations during a step, called the \textbf{Individual Gait Pattern} (IGP). In order to asses the stability of statistical tools developed to analyze the IGP a large volume of data is needed, which is proving to be complicated as conducting clinical trials can be lengthy and costly.

The issue of small sample size is a recurrent issue in health studies, moreover, sharing clinical data is challenging because de-identification methods, such as removing directly identifying variables (e.g., names, social security numbers), have proven insufficient for protecting personal data~\cite{b42,b43,b44}. 

Recently described by Wang et al. (2024) \cite{Wang24} as "artificial data that can mimic the statistical properties, patterns, and relationships observed in real-world data", synthetic data, first introduced by Rubin (1993)~\cite{b11} and further developed by Raghunathan et al. (2003) ~\cite{raghunathan2003} offers a successful solution. 

Synthetic gait data generation methods have been developed to improve our understanding of gait impairments. Some of these methods are based on visual representation of the human gait captured by cameras such as the \textit{VersatileGait} \cite{VersatileGait}, or the synthetic Parkinsonian Gait generated by Chavez et al. (2022) \cite{ParkisonGait}. Kim et al. (2023) developed a method to generate gait patterns characterized by joint angles time series \cite{Kim_Hargrove_2023}. Most of the gait generation methods require already big original dataset as they are machine-learning based and need to be trained on a bigger data set to perform well.

However, there is a significant gap in the literature regarding the generation of synthetic QTS data, which are nevertheless a very good choice to describe any joint movement during gait. n this article we propose two scientific contributions, which together form a complete solution for generating synthetic QTS describing gait data. The first contribution, described in \Cref{sec:framework} gives a comprehensive and flexible framework based on dimension reduction that transforms the data such that the individual information is contained in a score matrix while conserving the shape of the curves in forms of principal modes of variation. Thanks to this framework, any tabular synthetic data generation method can be used to generate QTS. The second contribution is the implementation of a tabular synthetic data generation method based on nearest neighbors weighting, aiming to generate some data with respect to the original geometry and with a high fidelity, that performs for small data sets and without any previous knowledge of the data. We also indicate how to use the method for private synthetic gait data.

In the second section of this article, we describe the clinical data as well as the key concepts to understand the IGP and QTS objects (see \Cref{sec:gaitdata}). The third section details the comprehensive framework to reduce unit QTS data to tabular data (see \Cref{sec:framework}) as well as the algorithm we propose to generate synthetic data: \texttt{SynGait} (see \Cref{sec:syngait}. We also give advice on how to tune the hyper-parameters of the method. The fourth section details two tabular synthetic data generation method (see \Cref{sec:compare_method}) to compare our method with, as well as metrics to evaluate the quality of the synthetic data it produces (see \Cref{sec:performance-metrics}).
The fifth section illustrates the method with the presentation of MS gait data and synthetic data generated from it, the results show that this method performs really well in terms of fidelity. Metrics show a good geometry preservation, while guarantying that the data is new and the variability of the synthetic data is preserved as well as possible (see \Cref{sec:results}). The results are discussed in \Cref{s:discussion}.

\section{Gait data}
\label{sec:gaitdata}
\subsection{How is Gait measured?}

Walking is defined by the International Classification of Functioning, Disability and Health of the World Health Organization as: "Moving along a surface on foot, step by step, so that one foot is always
on the ground, such as when strolling, sauntering, walking forwards, backwards, or sideways" \cite{ICF}. Gait is the manner in which human walk, its analysis is very useful in health diagnosis, rehabilitation or sports. 

Gait can be characterized using various numerical systems that can be grouped into three categories:
\begin{itemize}
    \item \textbf{Pressure sensor platforms and mats.} This approach uses sensors placed under a mat that measure the reaction force of the ground during a step. They provide quantitative information on walking cycles. \cite{PressureMat2017}
    \item \textbf{Image analysis systems.} This approach involves utilizing cameras to record subjects as they walk, then generates silhouettes or spatio-temporal parameters \cite{SilhouetteGait2006}. Subjects can also be wearing sensors to flag the position of joints or body parts and algorithms can then estimates 3D movements \cite{Vicon2017}.
    \item  \textbf{Wearable sensors.} method assesses gait kinematics or kinetics by placing multiple sensors (such as accelerometers, gyroscopes, and magnetometers) placed on human joints or limbs. Tao et al. (2012) provide explanations on the basic principles of motion sensors systems\cite{sensorWeijun2012}. This technique is often less costly and less invasive.  
    
\end{itemize}

The method we selected for determining gait patterns utilizes wearable sensors.

 The next two sections are dedicated to (i) detailing the clinical study from which the data has been collected (\Cref{sec:MYO}) and (ii) giving all necessary background on how the individual gait patterns have been measured and processed (\Cref{sec:IGP}).

\subsection{The ancillary MYO study}
\label{sec:MYO}
The MYO clinical study is a 2018 exploratory study led by Prof. D.A. Laplaud and P.A. Gourraud at the Nantes Teaching University Hospital. It aims at analyzing nerve signals of MS patients collected by an electronic wristband coined MYO. An amendment to the original protocol was approved so that we could make a total of 30 patients wear the eGait device and collect gait data. The inclusion period was September 2019 to May 2020. Each included patient was asked to perform the \textit{Timed 25-Foot Walk}~\cite{T25FW2012} (T25FW) test while wearing the eGait device. A total of 27 \textbf{individual gait pattern} (IGP) were successful obtained following a standardized custom-made pipeline including signal preprocessing steps, gait cycle segmentation and IGP computation. The overall disability of each patient was also assessed. To this end, the expanded disability status scale (EDSS) was used. The scale ranges from 0 (normal neurological examination) to 10 (death due to MS)~\cite{b34}.

\subsection{The individual gait pattern}
\label{sec:IGP}
\subsubsection{Data collection}

The eGait device is made of three elements:

\begin{enumerate}
    \item An inertial measurement unit (IMU): this is an electronic device composed of a 3-axis accelerometer, a 3-axis gyroscope and a 3-axis magnetometer, the data of which are fused together to calculate the orientation of the IMU with respect to a reference frame over time. The Mbientlab MetaMotionR (MMR) IMU was used for this study and was placed on the right hip of the subjects, at the level of the iliac crest.
    \item A smartphone with a dedicated mobile app: the smartphone connects to and sets up the IMU via bluetooth. Data is recorded on the phone in the form of a time series of consecutive orientation (hence, 3D rotations) of the IMU over time.
    \item A set of statistical methods: statistical methods dedicated to the analysis of rotation-valued time series or functional data were developed as part of the eGait device.
\end{enumerate}

Three-dimensional object orientation is nothing but a 3-dimensional rotation. Mathematically, this can be expressed in various form: 3-dimensional rotation matrix, Euler angles, Tait-Bryan angles, roll-pitch-yaw angles, axis-angle representation or unit quaternions. The orientation of the IMU is returned using the latter representation, which offers the best storage compression and avoid gimbal lock issues (some rotations are not uniquely defined when using alternative representations).

The eGait device therefore ultimately provides what we coined the IGP of an individual, which describes the right hip rotation over time during a typical gait cycle. It is expressed as a sequence of unit quaternions on a grid of 101 points, where each time points represents a percentage of the overall stride duration from $0\%$ to $100\%$ with a step of $1\%$~\cite{b30}. In the next section, we will briefly summarize key concepts about unit quaternions that will prove useful in the remainder of the article.

\subsubsection{Unit quaternion time series}
\label{s:uqts}

The space $\mathbb{H}$ of quaternions is isomorphic to $\mathbb{R}^4$ and an element $\mathbf{q} \in \mathbb{H}$ is an hypercomplex number that can be viewed as an extension of complex numbers, which reads:

\begin{equation}
\label{eq:quat}
\mathbf{q} := (q_w, q_x, q_y, q_z)^\top \equiv q_w + \mathbf{i} q_x + \mathbf{j} q_y + \mathbf{k} q_z \in \mathbb{R}^4,
\end{equation}
where $\mathbf{i}$, $\mathbf{j}$ and $\mathbf{k}$ follow the rule $\mathbf{i}^2=\mathbf{j}^2=\mathbf{k}^2=\mathbf{ijk}=-1$. Quaternions have been formalized by Sir William Hamilton in 1943 and their properties can be found in many textbooks~\cite{b16}. In particular, $\mathbb{H}$ is a 4-dimensional associative normed division algebra over the real numbers.

\begin{definition}[unit quaternion]
A \textit{unit quaternion} is a quaternion $\mathbf{q} \in \mathbb{H}$ -- hence satisfying \cref{eq:quat} -- of unit norm, \textit{i.e.} such that $\| \mathbf{q} \|^2 = q_w^2 + q_x^2 + q_y^2 + q_z^2= 1$.
\end{definition}

The space $\mathbb{H}_u$ of unit quaternions is a \textit{Lie group} isomorphic to the special unitary group $\mathrm{SU}(2)$, which is a double coverage of the space $\mathrm{SO}(3)$ of 3-dimensional rotation matrices~\cite{b32}. Lie groups are both groups and differentiable Riemaniann manifolds which guarantees existence of a tangent space at each point. The fact that unit quaternions double-cover 3-dimensional rotations translates into the fact that $\mathbf{q}$ and $-\mathbf{q}$ encode the same rotation. There is a link between the axis-angle representation $(\mathbf{u}, \theta) \in \mathbb{S}^2 \times \mathbb{R}$ of a 3-dimensional rotation and its unit quaternion representation, which reads:

\begin{equation}
\label{eq:quat_rot}
\mathbf{q} = \cos \frac{\theta}{2} + \sin \frac{\theta}{2} \left( u_x \mathbf{i} + u_y \mathbf{j} + u_z \mathbf{k} \right).
\end{equation}

We can now define a mathematical structure to host the data produced by the eGait device.

\begin{definition}[unit quaternion time series]
\label{def:qts}
    A \textit{unit quaternion time series} (QTS) $\mathbf{Q}$ is a sequence of unit quaternions along a time grid of ordered points $t_1 < \dots < t_p$ such that $\mathbf{Q}(t_k) = \mathbf{q}_k \in \mathbb{H}_u$, for $k \in [\![ 1, P ]\!]$.
\end{definition}

The geometry, topology and algebra of $\mathbb{H}_u$ does not resemble at all usual Euclidean vector spaces into which classical statistical methods apply. In the following section, we will detail a comprehensive framework for performing statistical analysis for QTS.

\section{Proposed Method}
\label{sub:SDG}

The main objective of the proposed approach is to create synthetic QTS data that belong to the same unknown manifold as the original QTS data. This is achieved by two original contributions. 

First, we propose a comprehensive and flexible framework for synthesizing unit QTS data that relies on functional principal component analysis (fPCA)~\cite{b9} which provides \textit{principal components} in the form of functions that capture the principal modes of variation of a functional data set. The \textit{scores} of a functional datum are then defined as the projection of the centered functional datum onto the principal functions, \textit{i.e.} the inner product between the centered functional datum and each principal function. The key idea is to use fPCA to transform the original functional data set $v_1, \dots, v_n$ into a set of $n-1$ principal functions $\phi_1, \dots, \phi_{n-1}$ and a more conventional tabular data set in the form of a numeric matrix of shape $n \times (n-1)$ storing the functional scores. Synthesized versions of the original functional data set are subsequently obtained by first synthesizing the scores and then recombining them with the original principal functions kept untouched to generate synthetic log-quaternion functional data. This first key contribution is detailed in \Cref{sec:syngait} and summarized in \Cref{fig:syngait-overview}.

The second contribution focuses on how we actually synthesize the functional score matrix. While there is a plethora of methods developed in the literature for the synthesis of tabular data, we argue that they have hard time in producing data that belong to the unknown manifold onto which the original data belongs. We therefore propose a generalization of the avatar method introduced in~\cite{avatar} to achieve the synthesis of the functional scores, this flexible avatar method is detailed in \Cref{sec:syngait}.

\subsection{From time series to functional data analysis}
\label{sec:framework}
\subsubsection{From QTS to functions in \texorpdfstring{$\mathbb{R}^3$}{R3}}
\label{s:tangent}

The space of unit quaternions is topologically the 3-sphere $\mathbb{S}^3$ which is a Lie group, \textit{i.e.} both a group and a differentiable manifold. This means that, in each point of the space, there is existence of a tangent space isomorphic to $\mathbb{R}^3$. The interested user can refer to \cite{sola2018micro} for a micro-theory of Lie group with special focus on some particularly useful groups such as $\mathbb{S}^3$. In particular, the Lie algebra is the tangent space of a Lie group in its neutral element, here $\mathbf{e} = (1,0,0,0)^\top$. 

\begin{definition}[logarithmic map]
\label{def:log}
Any unit quaternion $\mathbf{q} \in \mathbb{H}_u$ can be mapped into the Lie algebra via the following logarithmic map:
$$
\begin{array}{rccl}
\log : & \mathbb{H}_u & \to &  \mathcal{T}_\mathbf{e} \mathbb{H}_u \cong \mathbb{R}^3 \\
& \mathbf{q} = (q_w, q_x, q_y, q_z)^\top & \mapsto & \frac{\acos q_w}{\sqrt{q_x^2 + q_y^2 + q_z^2}} (q_x, q_y, q_z)^\top
\end{array}
$$
If one uses the rotation parametrization of a unit quaternion as described in Eq. \eqref{eq:quat_rot}, the mapping boils down to $\log \mathbf{q} = \frac{\theta}{2} \mathbf{u}$.
\end{definition}

Conversely, the inverse logarithmic map also exist and maps from the Lie algebra back to the space of unit quaternions.

\begin{definition}[exponential map]
\label{def:exp}
Any vector in $\mathcal{T}_\mathbf{e} \mathbb{H}_u \cong \mathbb{R}^3$ can be mapped into a unit quaternion $\mathbf{q} \in \mathbb{H}_u$ via the following exponential map:

$$
\begin{array}{rccl}
\exp : & \mathcal{T}_\mathbf{e} \mathbb{H}_u \cong \mathbb{R}^3 & \to &\mathbb{H}_u \\
& \mathbf{v} = (v_x, v_y, v_z)^\top & \mapsto & \cos \| \mathbf{v} \| + \frac{\sin \| \mathbf{v} \| }{\| \mathbf{v} \|} \left( v_x \mathbf{i} + v_y \mathbf{j} + v_z \mathbf{k} \right).
\end{array}
$$
\end{definition}

From a unit quaternion time series as in \Cref{def:qts}, we can therefore introduce the corresponding log-quaternion time series.

\begin{definition}[log-quaternion time series]
\label{def:logqts}
Given a unit QTS $\mathbf{Q}$ observed on a time grid of ordered points $t_1 < \dots < t_p$ such that $\mathbf{Q}(t_k) = \mathbf{q}_k \in \mathbb{H}_u$, for $k \in [\![ 1, P ]\!]$, we define its \textit{log-quaternion time series} (log-QTS) $\mathbf{V}$ such that $\mathbf{V}(t_k) = \mathbf{v}_k = \log \mathbf{q}_k \in \mathcal{T}_\mathbf{e} \mathbb{H}_u \cong \mathbb{R}^3$, for $k \in [\![ 1, P ]\!]$.
\end{definition}

Lastly, we take on the view of functional data~\cite{b40} rather than time series analysis and therefore define a functional representation of a log-QTS using cubic B-spline interpolation.

\begin{definition}[log-quaternion functional datum]
\label{def:logqfd}
Given a log-QTS $\mathbf{V}$ from \Cref{def:logqts} observed on a time grid of ordered points $t_1 < \dots < t_p$ such that $\mathbf{V}(t_k) = \log \mathbf{q}_k \in \mathcal{T}_\mathbf{e} \mathbb{H}_u \cong \mathbb{R}^3$, for $k \in [\![ 1, P ]\!]$, we define its corresponding \textit{log-quaternion functional datum} (log-QFD) as the cubic B-spline interpolation of the observed points in $\mathbf{V}$.
\end{definition}

In the next section, we will explain in details a novel method for synthesizing QTS from \Cref{def:qts}, using, under the hood, the log-QFD representation from \Cref{def:logqfd}.

\subsubsection{Functional principal component analysis}
\label{sec:fpca}

This section gives the necessary background for understanding functional PCA and therefore is based on a generic sample $x_1, \dots, x_n$ of $n$ independent and identically distributed random functions. Let us assume that they belong to $L^2 \left( \mathcal{I}, \mathbb{R}^d \right)$, where $\mathcal{I} \subseteq \mathbb{R}$. When $d = 1$, we usually say that we are dealing with univariate functional data while, when $d > 1$, we say that we are dealing with multivariate functional data. Standard PCA~\cite{b7,b8} can be extended to such random functions, giving its name to functional PCA~\cite{b9}. This requires to define functional versions of the sample mean and sample covariance matrix.

\begin{definition}[sample mean]
We define the \textit{sample mean} of a functional data set $x_1, \dots, x_n$ as the random function $\overline{x}_n$ which reads:
$$
\overline{x}_n := \frac{1}{n} \sum_{i=1}^n x_i.
$$
\end{definition}

The concept of covariance matrix in functional spaces translates into a covariance operator, defined through a covariance kernel function.

\begin{definition}[sample covariance kernel function]
We define the \textit{sample covariance kernel function} as the random function $\widehat{\sigma}$ which reads:
$$
\begin{array}{rccl}
\widehat{\sigma}: & \mathcal{I} \times \mathcal{I} & \to & \mathcal{M}_d(\mathbb{R}) \\
& (s, t) & \mapsto & \frac{1}{n - 1} \sum_{i=1}^n \left( x_i(t) - \overline{x}_n(t) \right) \left( x_i(s) - \overline{x}_n(s) \right)^\top.
\end{array}
$$
\end{definition}

The sample covariance kernel function can then be integrated out to obtain the covariance operator.

\begin{definition}[sample covariance operator]
We define the \textit{sample covariance operator} as the random function $\widehat{V}$ which reads:
$$
\begin{array}{rccl}
\widehat{V}: & L^2(\mathcal{I}, \mathbb{R}^d) & \to & L^2(\mathcal{I}, \mathbb{R}^d) \\
& f & \mapsto & \begin{array}{rccl}
\widehat{V}(f): & \mathcal{I} & \to & \mathbb{R}^d \\
& t & \mapsto & \int_\mathcal{I} \widehat{\sigma}(t,s) f(s) ds.
\end{array}
\end{array}
$$
\end{definition}

The principal components $\phi_1, \dots, \phi_{n-1}$ correspond to the eigenfunctions of $\widehat{V}$ associated to the $n - 1$ non-null eigenvalues $\lambda_1, \dots, \lambda_{n-1}$. For a given functional datum $x_i$ in the functional data set $x_1, \dots, x_n$, we define the functional scores:

$$
f_{ik} = \left \langle x_i - \overline{x}_n, \phi_k \right \rangle = \sum_{j=1}^d \int_\mathcal{I} \left( x_i^{(j)}(t) - \overline{x}_n^{(j)}(t) \right) \phi_k^{(j)}(t) dt  , \quad k \in [\![1, n-1]\!].
$$

The (multivariate) functional PCA therefore provides a set $\phi_1, \dots, \phi_{n-1}$ of principal functions along with a functional score matrix $\mathbb{F}$ of shape $n \times (n-1)$. In practice, we use the approach described in~\cite{b38} which is implemented in the \texttt{R} package \texttt{MFPCA}~\cite{b14}.

\subsection{Overview of the proposed method}

We start with an original sample $\Quat_1, \dots, \Quat_n$ of $n$ unit QTS observed on the same time grid $t_1 < \dots < t_p$ such that $\Quat_i(t_k) = \quat_{ik} \in \mathbb{H}_u$. 

\begin{figure*}[!htpb]
\centering
\includegraphics[width=\textwidth,page=2]{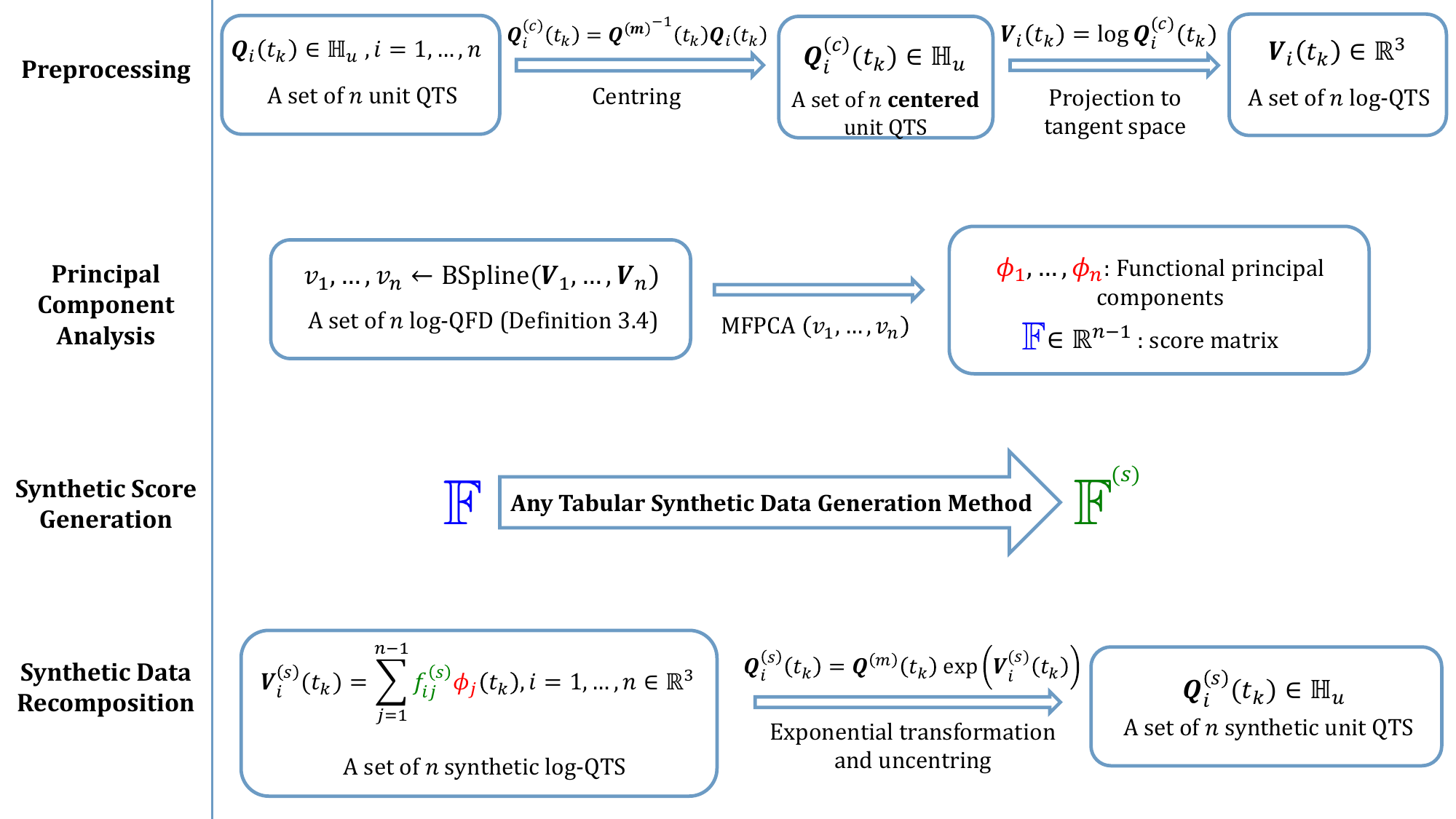}
\caption{\textbf{\texttt{SynGait.}} Schematic overview of the proposed comprehensive framework for unit QTS synthetic data generation.}
\label{fig:syngait-overview}
\end{figure*}

Hereafter, we detail every step of the synthetic gait data generation comprehensive framework that we propose. Steps are illustrated in \Cref{fig:syngait-overview}.

\begin{description}
    \item[Centering.] As in standard PCA, we start by centering the data around the pointwise mean. At each time point $t_k$ of the initial grid of observation, we compute the Fréchet mean of the $n$ unit quaternions $\quat_{1k}, \dots, \quat_{nk}$ associated to the geodesic distance between two unit quaternions, which reads:
    \begin{equation}
    \label{eq:quat-mean}
    \quat_k^{(m)} := \Quat^{(m)}(t_k) = \underset{ \quat \in \mathbb{H}_u }{\arg \min}  \sum_{i=1}^{n} d_g^2(\quat_{ik}, \quat) \in \mathbb{H}_u, \quad k \in [\![1, p]\!],
    \end{equation}
    where $d_g(\quat_1,\quat_2):=\left\| \log(\quat^{-1}_1 \quat_2) \right\|$. This defines the mean QTS $\Quat^{(m)}$ which we then use the centered QTS $\Quat_1^{(c)}, \dots, \Quat_n^{(c)}$ as:
    \begin{equation}
    \label{eq:qts-centered}
    \quat_{ik}^{(c)} := \Quat_i^{(c)}(t_k) = \left( \quat_k^{(m)} \right)^{-1} \quat_{ik} \in \mathbb{H}_u, \quad k \in [\![1, p]\!], \quad i \in [\![1, n]\!].
    \end{equation}

    \item[Projection to tangent space.] Next, we project the centered QTS which take values in the Lie group $\mathbb{H}_u$ into the corresponding Lie algebra by defining the log-QTS taking values in $\mathbb{R}^3$ out of the centered QTS as:
    \begin{equation}
    \label{eq:qts-log}
    \mathbf{v}_{ik} := \mathbf{V}_i(t_k) = \log \Quat_i^{(c)}(t_k) \in \mathbb{R}^3, \quad k \in [\![1, p]\!], \quad i \in [\![1, n]\!].
    \end{equation}

    \item[Multivariate functional PCA.] Next, we transform the log-QTS data set $\mathbf{V}_1, \dots, \mathbf{V}_n$ into a functional data set $v_1, \dots, v_n$ using univariate cubic B-splines and perform a multivariate functional PCA as described in \Cref{sec:fpca} to compute the principal components $\phi_1, \dots, \phi_{n-1}$ and the associated score matrix $\mathbb{F}$ of shape $n \times (n - 1)$. 

    \item[Synthetic score generation.] Next, we generate synthetic scores using any synthetic data generation method designed for tabular data. A good option for data living on an unknown manifold is the flexible avatar method for synthetic tabular data generation presented in \Cref{sec:syngait}.

    \item[Unit QTS synthesis.] We finally produce synthetic unit QTS by (i) combining the generated scores with the original principal components, (ii) evaluating the resulting functions on the original time grid $t_1 < \dots < t_p$, (iii) mapping back the resulting log-QTS to the space of unit QTS and (iv) adding back the mean QTS:
    \begin{equation}
    \label{eq:qts-syn}
    \Quat_i^{(s)}(t_k) = \Quat^{(m)}(t_k) \exp \left( \sum_{j = 1}^{n-1} f_{ij}^{(s)} \phi_j(t_k) \right) \in \mathbb{H}_u,
    \end{equation}
    for $k \in [\![1, p]\!]$ and $i \in [\![1, n]\!]$. 
\end{description}

\subsection{A more flexible avatar method for synthetic tabular data generation}
\label{sec:syngait}

The avatar method is a patient-centered synthetic data generation method. It uses each observation to create a simulation in a feature space leading to the creation of a single synthetic observation~\cite{avatar}. Originally developed for anonymization purposes, we propose a more flexible adaptation of this method by using the Dirichlet distribution instead of the exponential distribution as originally proposed. This flexible approach permits the user to select a concentration parameter for the weight distribution, thereby having an impact on the variability and on the privacy of the generated synthetic data set. The three main steps of the method are described below.

\begin{description}    

    \item[Nearest neighbor search.] First, we compute the pairwise Euclidean distance matrix $\mathbb{D}$ between rows of the score matrix $\mathbb{F}$ reduced to its first $\tau$ columns. The input parameter $\tau$ therefore determines the amount of dimensionality reduction used to find the nearest neighbors. For each data point, we subsequently conduct a search of its $\gamma$ nearest neighbors using this distance matrix. This leads to $n$ vectors $\boldsymbol{\ell}_1, \dots, \boldsymbol{\ell}_n$ of size $\gamma$ containing the indices of the $\gamma$ nearest neighbors of each observation and $n$ vectors $\mathbf{d}_1, \dots, \mathbf{d}_n$ containing the corresponding distances.

    \item[Nearest neighbor weights.] We next generate weights for each neighbor of each observation. These weights belong to the $\gamma$-simplex and we therefore need to resort to a statistical distribution with such a support. The Dirichlet distribution is a natural choice~\cite{b37}. For the $i$-\textit{th} observation, this distribution has $\gamma$ concentration parameters $\alpha_{i1}, \dots, \alpha_{i\gamma}$, which can be compressed into a $\gamma$-dimensional vector $\boldsymbol{\alpha}_i$. We propose to make the concentration parameters dependent upon the distance between the $i$-\textit{th} observation and its $\gamma$ nearest neighbors as follows:
    \begin{equation}
    \label{eq:alphas}
    \boldsymbol{\alpha}_i := \frac{\alpha_0}{\sum_{j=1}^\gamma d_{ij}^{-1}} \mathbf{d}_i^{-1} \in \left( \mathbb{R}^{+\star} \right)^{\gamma}, \quad i \in [\![1, n]\!].
    \end{equation}
    where $\alpha_0$ is the sum of the concentration parameters and acts as a gauge of variability on the generated weights. The output of this step is a collection of $n$ vectors $\mathbf{w}_1, \dots, \mathbf{w}_n$, such that $\mathbf{w}_i \sim \mathrm{Dir}(\boldsymbol{\alpha}_i)$, containing the weights associated to each nearest neighbor of each observation.

    \item[Functional score synthesis.] We then produce synthetic scores for observation $i$ as a weighed average of the scores of its $\gamma$ nearest neighbors:
    \begin{equation}
    \label{eq:scores-syn}
    \mathbf{f}_i^{(s)} = \sum_{j = 1}^\gamma w_{ij} \mathbf{f}_{\ell_{ij}} \in \mathbb{R}^{n-1}, \quad i \in [\![1, n]\!].
    \end{equation}
\end{description}

The following \Cref{alg:SQTSG} offers a summarized pseudoalgorithmic view of the proposed method with both comprehensive framework and synthetic score generation methods combined.

\begin{algorithm}
\caption{Synthetic Gait Data Generation (\texttt{SynGait})}
\label{alg:SQTSG}
\begin{algorithmic}[1]
\Procedure{SynGait}{$\Quat_1, \dots, \Quat_n$, $\gamma$, $\tau$, $\alpha_0$} 

    -----------------------------------------------------------------------------------
    
    \State \textbf{Inputs:}
    \State $\quad \Quat_1, \dots, \Quat_n$ such that $\Quat_i(t_k) = \quat_{ik} \in \mathbb{H}_u$: A set of $n$ unit QTS;
    \State $\quad \gamma \in [\![1,n]\!]$: The number of nearest neighbors;
    \State $\quad \tau \le n - 1$: The number of principal components to keep for the nearest neighbor search;
    \State $\quad \alpha_0 > 0$: The sum of the concentration parameters of the Dirichlet distribution used to sample weights associated to each neighbors.

    \State \textbf{Output:}
    \State $\quad \Quat_1^{(s)}, \dots, \Quat_n^{(s)}$: A set of $n$ unit QTS synthesized from the original input set.

    -----------------------------------------------------------------------------------
    
    \State Compute the mean QTS $\Quat^{(m)}$ via Eq. \eqref{eq:quat-mean};
    \State Compute the set $\Quat_1^{(c)}, \dots, \Quat_n^{(c)}$ of centered unit QTS via Eq. \eqref{eq:qts-centered};
    \State Compute the set $\mathbf{V}_1, \dots, \mathbf{V}_n$ of log-QTS via Eq. \eqref{eq:qts-log};
    \State $v_1, \dots, v_n \gets$ BSpline($\mathbf{V}_1, \dots, \mathbf{V}_n$);
    \State $\left( \phi_1, \dots, \phi_{n-1}, \mathbb{F} \right) \gets$ MFPCA($v_1, \dots, v_n$);
    
    \For{$i=1,\dots,n$}
        \State Find the $\gamma$ nearest neighbors $\boldsymbol{\ell}_i$ with associated Euclidean distances $\mathbf{d}_i$ on the first $\tau$ components;
        \State Compute concentration parameters $\boldsymbol{\alpha}_i$ via Eq. \eqref{eq:alphas} from $\alpha_0$, $\mathbf{d}_i$;
        \State Sample $\gamma$ weights $\mathbf{w}_i \sim \mathrm{Dir}(\boldsymbol{\alpha}_i)$;
        \State Synthesize new scores $\mathbf{f}_i^{(s)}$ via Eq. \eqref{eq:scores-syn};
        \State Synthesize new unit QTS via Eq. \eqref{eq:qts-syn}.
    \EndFor
\EndProcedure
\end{algorithmic}
\end{algorithm}

\Cref{alg:SQTSG} is implemented in the \texttt{R} package \textbf{egait-sd}, which is currently hosted on a private repository. We plan to make it public in the near future.

As one can notice, there are two types of inputs to the algorithm: (i) the input data and (ii) a set of three hyper-parameters ($\gamma$, $\tau$ and $\alpha_0$) which play a critical role on the properties of the set of synthetic unit QTS that the method can produce. The next section outlines a method for tuning these parameters.

\subsection{Optimization of the hyper-parameters}
\label{sec:hpo}

The choice of the number $\gamma$ of nearest neighbors to use for score synthesis is a trade-off between generating a set of unit QTS whose shapes are similar to those from the original set thus keeping a maximal variance (small $\gamma$) and generating a set of unit QTS with more anonymized synthetic data (high $\gamma$). 

The choice of the number $\tau$ of principal components to keep for searching the nearest neighbors should be made by considering the cumulative percentages of inertia explained by the principal components. Searching for the nearest neighbors using a number of principal components that only explain a small fraction of inertia could result in a bigger inertia among the synthetic data set but could also result in synthetic data almost identical to the original data.

Finally, the choice of the total concentration $\alpha_0$ for the Dirichlet distribution acts as a trade-off between completely deterministic weights on the nearest neighbors ($\alpha_0 \to \infty$) which yields perfect anonymization and random selection of one of the nearest neighbors ($\alpha_0 \to 0$) which forbids anonymization.

We propose a method to help users select these parameters that takes as input a data set of unit QTS and returns the best parameter combinations according to two considerations:

\begin{description}
    \item[Minimum distance criterion.] First, the minimum distance $d_{\min}$ between any two synthetic scores $\mathbf{f}_i^{(s)}$ and $\mathbf{f}_{i^\prime}^{(s)}$ and between any two original and synthetic scores $\mathbf{f}_i$ and $\mathbf{f}_{i^\prime}^{(s)}$, which reads:
    \begin{equation}
    \label{eq:dmin}
    d_{\min} := \min \left( \min_{(i, i^\prime) \in [\![1,n]\!], i \ne i^\prime} d \left( \mathbf{f}_i^{(s)}, \mathbf{f}_{i^\prime}^{(s)} \right), \min_{(i, i^\prime) \in [\![1,n]\!]} d \left( \mathbf{f}_i, \mathbf{f}_{i^\prime}^{(s)} \right) \right),
    \end{equation}
    is chosen by the user. We recommend using a value at least as big as one tenth of the smallest distance between observations in the original data set. This constraint helps in guaranteeing that synthetic data is both new (as opposed to a very small variation of the original data) and not too close from one another.
    \item[Maximum distance criterion.] Then, the maximum distance $d_{\max}$ between any two synthetic scores $\mathbf{f}_i^{(s)}$ and $\mathbf{f}_{i^\prime}^{(s)}$, which reads:
    \begin{equation}
    \label{eq:dmax}
    d_{\max} := \max_{(i, i^\prime) \in [\![1,n]\!], i \ne i^\prime} d \left( \mathbf{f}_i^{(s)}, \mathbf{f}_{i^\prime}^{(s)} \right),
    \end{equation}
    should be as big as possible. This helps to produce synthetic data whose variance gets as close as possible to the one from the original data.
\end{description}

Those distances are computed for each parameter combination that the user wants to try, and each combination is repeated at least 10 times (the amount of repetition is user-defined). There is a dedicated function in the \texttt{R} package \textbf{egait-sd} which returns a table with all parameter combinations ranked in decreasing order of the average (over repetitions) maximum distances, as well as graphs that helps understanding the relationship between the parameters and their impact on the distances. Note that if the synthetic data is generated for anonymization purposes, one should choose a high number of neighbors and a higher threshold for $d_{\min}$.

\section{Comparison to existing synthetic tabular data generation methods}
\subsection{Compared methods}
\label{sec:compare_method}
The \texttt{SynGait} method proposed in \Cref{sub:SDG} features an outstanding ability to preserve the geometry of the original data in the generated synthetic data sets. Nonetheless, at the core of the method, it boils down to generating synthetic tabular data (the functional score matrix). As such, one could think of using any of the existing methods in the literature that are dedicated to this purpose.

The \href{https://docs.sdv.dev/sdv}{Synthetic Data Vault} (SDV) is a system of open-source libraries developed by the \texttt{datacebo} company~\cite{SDV}. It allows users to generate and evaluate synthetic data for single table, multi-table and time-series data~\cite{b56}. State-of-the-art methods for synthetic tabular data generation nowadays revolve around two main categories: (i) copula-based methods and (ii) methods based on generative adversarial network (GAN) models. We therefore chose to compare our nearest-neighbor weighting approach to reference algorithms in the SDV (with default parameters) that implement a copula-based and a GAN-based approach (CTGAN) and we used a number of performance metrics to carry out such a comparison. The next two subsections give some details about the compared methods while \Cref{sec:performance-metrics} defines the performance metrics that we will use for the comparison.

\subsubsection{Copula method}

Copulas have been extensively used in the literature to generate synthetic data by estimating the joint distribution and generating new data from it \cite{b61,b62}. Copulas have been introduced in~\cite{sklar1959fonctions} in which the following founding theorem can be found:

\begin{theorem}[Sklar's theorem]
\label{thm-sklar}
Let $F \in \mathcal{F}(F_1, \dots, F_p)$ be a p-dimensional distribution function with marginals $F_1, \dots, F_p$. Then there exist a copula $C$ which is a p-dimensional distribution function on $[0,1]^p $ with uniform marginals such that 
\begin{equation}
\label{eq-sklar}
F(x_1, \dots, x_p) = C\left(F_1(x_1), \dots, F_p(x_p)\right), \quad x_j \in \mathbb{R}, \quad j \in [\![1,p]\!].
\end{equation}
\end{theorem}

Substantially, it means that one can approximate any $p$-dimensional distribution function by estimating a distribution function $C$, termed the \textit{copula}, on $[0,1]^p$ with uniform marginals. Estimators of $C$ can be parametric or non-parametric. A standard approach is to rely on the Gaussian distribution function as in~\cite{meyer2021copula}. Specifically, they estimate the copula function $C$ for a $p$-variate random variable $\mathbf{X}$ from an $n$-sample $\mathbf{X}_1, \dots, \mathbf{X}_n \sim \mathbf{X}$, using the following steps:
\begin{itemize}
\item[(i)] Estimate the marginal distribution functions $\hat{F}_1, \dots, \hat{F}_p$ by the corresponding empirical cumulative distribution functions (CDF): 
\begin{equation*}
\widehat{F}_j(x) = \frac{1}{n} \sum_{i=1}^n \vmathbb{1} \left(X_{ij} \leq x \right);
\end{equation*}
\item[(ii)] Define the pseudo-observations $\widehat{U}_{ij} := \widehat{F}_j(X_{ij})$, which makes the variables $\widehat{U}_{ij}$ close to being sampled from a uniform distribution (it would be exactly the case if the marginal distribution functions $F_j$'s had been the true ones and not the empirical CDFs);
\item[(iii)] Define $Z_{ij} := \Phi^{-1} \left( \widehat{U}_{ij} \right)$, where $\Phi$ is the cumulative distribution function of the standard normal distribution, which makes the variables $Z_{ij}$ close to being sampled from a standard normal distribution;
\item[(iv)] Estimate the correlation matrix from the sample $\mathbf{Z}_1, \dots, \mathbf{Z}_n$ as:
\begin{equation*}
\widehat{\Sigma} = \frac{1}{n} \sum_{i=1}^n \mathbf{Z}_i \mathbf{Z}_i^\top.
\end{equation*}
\item[(v)] For any $p$-tuple $(x_1, \dots, x_p) \in [0,1]^p$, define the estimator $\widehat{C}$ of the copula function $C$ as the distribution function of a centered $p$-dimensional Gaussian distribution with covariance matrix $\widehat{\Sigma}$ applied to $(\Phi^{-1}(x_1), \dots, \Phi^{-1}(x_p))$ which reads:
\begin{equation*}
    \widehat{C}(x_1, \dots, x_p) := \Phi_{\widehat{\Sigma}} \left( \Phi^{-1}(x_1), \dots, \Phi^{-1}(x_p) \right),
\end{equation*}
where $\Phi$ is the CDF of the standard normal distribution and $\Phi_{\widehat{\Sigma}}$ is the CDF of the centered multivariate normal distribution with covariance matrix $\widehat{\Sigma}$.
\end{itemize}

This effectively provides an estimator of the copula function which makes it the CDF of a multivariate Gaussian distribution which it is easy to sample from by (i) performing the Cholesky decomposition of $\widehat{\Sigma} = \mathbb{L} \mathbb{L}^\top$, (ii) sampling independently $p$ values from the standard normal distribution, (iii) put them in a $p$-dimensional vector and (iv) pre-multiply this vector by $\mathbb{L}$. The values $y_1, \dots, y_p$ in the resulting vector are sampled from $\Phi_{\widehat{\Sigma}}$.

Using \Cref{thm-sklar}, we have an approximation of the CDF of the original data through:
\begin{equation}
\label{eq:sklar}
\widehat{F}(x_1, \dots, x_p) = \Phi_{\widehat{\Sigma}} \left( \Phi^{-1}\left( \widehat{F}_1(x_1) \right), \dots, \Phi^{-1} \left( \widehat{F}_p(x_p) \right) \right).
\end{equation}
Hence, once one have sampled the values $y_1, \dots, y_p$ from $\Phi_{\widehat{\Sigma}}$, we obtain a sample $x_1^{(s)}, \dots, x_p^{(s)}$ approximately from the distribution of the original data as $x_j^{(s)} = \widehat{F}_j^{-1}(\Phi(y_j))$, for $j = 1, \dots, p$, which essentially reverts \cref{eq:sklar}. The interested reader can refer to~\cite{SDV} for detailed description of the algorithms behind copula function estimation in the SDV. 

\subsubsection{Conditional tabular generative adversarial network (CTGAN) method}

Generative adversarial networks, or GANs, are machine learning frameworks which rely on a generative model $G$ that learns how to generate plausible data (in the sense that it resembles a target original data set) from data sampled from any pre-specified model (usually the Gaussian model) and a discriminating model $D$ that learns to distinguish between the target real data and the artificial data produced by the generator~\cite{b52}. The \textit{discriminator function}
\begin{equation}
\label{eq:gan-discriminator}
\begin{array}{rccl}
D_\mathbf{x} : & \Theta_d & \to & [0,1] \\
& \theta_d & \mapsto & D(\theta_d; \mathbf{x})
\end{array}
\end{equation}
is a differentiable function represented by a neural network which gets some data $\mathbf{x}$ either from the target real data or from the artificial one produced by the generator and outputs the probability of being sampled from the same distribution that generated the target real data. The \textit{generator function}
\begin{equation}
\label{eq:gan-generator}
\begin{array}{rccl}
G_\mathbf{z} : & \Theta_g & \to & \mathcal{X} \\
& \theta_g & \mapsto & G(\theta_g; \mathbf{z})
\end{array}
\end{equation}
maps a random input vector $\mathbf{z}$ from a latent space to a synthetic data sample that resembles real data.

\begin{figure}[!htpb]
\centering
\includegraphics[width=0.85\linewidth]{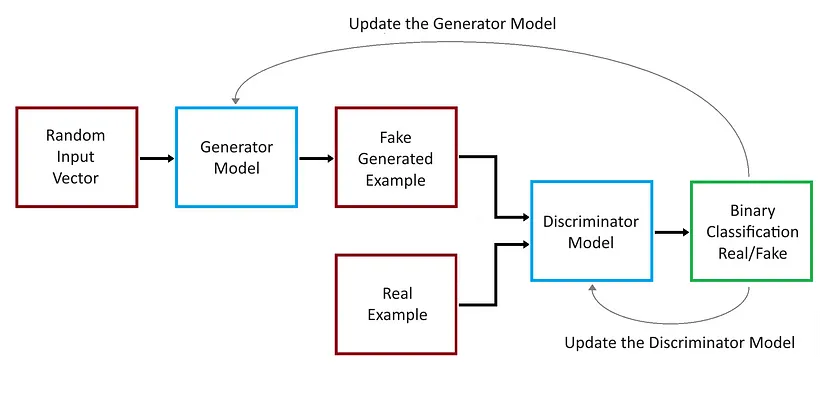}
\caption{Overview of how a GAN model is trained.}
\label{fig:GAN-principle}
\end{figure}

The parameters $\theta_d$ and $\theta_g$ are the weights of the networks that are learned during training. \Cref{fig:GAN-principle} provides an overview of the usual way GANs are trained\footnote{https://medium.com/@marcodelpra/generative-adversarial-networks-dba10e1b4424}.
Specifically, the generator model is updated iteratively until the discriminator cannot distinguish real from fake data. The generator has no direct access to real data, so in practice the discriminator is usually trained first then the generator is trained by trying to fool the discriminator. You can find more about how GANs are trained in~\cite{GAN_overview2018}. The architecture of the specific model for conditional tabular GANs (CTGANs) developed by \texttt{datacebo} is detailed in~\cite{xu2019modeling}. It integrates a conditional generator to resample efficiently to account for the imbalance in the categorical columns.

\subsection{Performance metrics}
\label{sec:performance-metrics}

Synthetic data are often evaluated according to three main criteria:

\begin{description}
    \item[Fidelity.] Measures to which point the synthetic data shares the same characteristics as the original one (mean, correlation, auto-correlation, visual inspections, etc.);
    \item [Utility.] Measures how different the results of the desired statistical analysis are when conducted on the original or synthetic data;
    \item [Privacy.] Measures how hard it is to identify real individuals from synthetic ones.
\end{description}

In the remainder of the paper, we mainly focus on fidelity measures as well as one privacy metric. All of these metrics are computed on the functional score matrix $\mathbb{F}$ (original and synthetic).

First we introduce a metric aiming to show how well the geometry is preserved thanks to some computational geometry.

\subsubsection{Frobenius distance on k-nearest neighbors graphs}
\label{sec:fb}
A good way to approximate the geometry of the unknown manifold $\mathbb{M}$ onto which a point cloud of size $n$ belongs is to build its $k$-nearest neighbors graph ($k$-NNG). In effect, $k$-NNGs are often used in computational geometry because there is an optimal value $k_n^\star$ that makes the $k_n^\star$-NNG close to $\mathbb{M}$~\cite{ComputationalGeom2012}.

\begin{figure}[!htpb]
\centering
\includegraphics[width=0.95\linewidth]{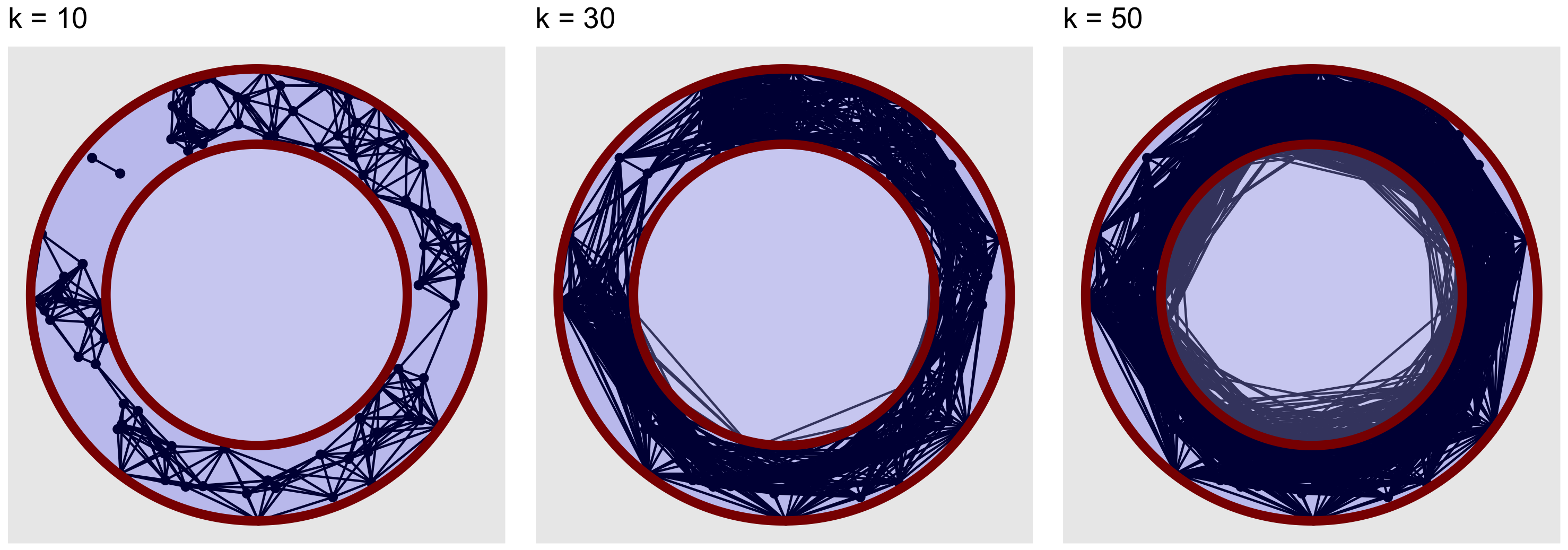}
\caption{\textbf{Manifold approximation via k-NNG.} A cloud of $n = 100$ points sampled in the space between the two red circles. Three k-NNGs are used to approximate the corresponding manifold: with $k=10$ (left panel), with $k=30$ (middle panel) and with $k=50$ (right panel). It illustrates that a value of $k$ close to $30$ is best in this case to approximate the space between the red circles.}
\label{fig:nng-principle}
\end{figure}

Since the manifold onto which the functional scores belong is unknown, there is no way to know the optimal value of $k$. Hence, we computed the k-NNG from the functional scores of both the original and synthetic data for all possible values of $k=1, \dots, n-1$. In practice, we used the \texttt{nng()} function from the \texttt{R} package \textbf{cccg}~\cite{b23} to compute undirected k-NNGs.

Next, for each value of $k=1,\dots,n-1$, we calculated the Frobenius distance between the adjacency matrices of the two k-NNGs as a measure of how close is the manifold onto which the synthetic data belongs to the manifold onto which the original data belongs. Specifically, we minimized the Frobenius distance over all possible permutations of the nodes since the manifold approximation via k-NNGs should not depend upon the ordering of the points.

\subsubsection{The RV coefficient}

Another interesting metric that quantifies the squared correlation between two data matrices observed on the same individuals is the rho-vector (RV) coefficient~\cite{josse2008testing,b35}. We compute this coefficient to measure the degree of association between the original score matrix $\mathbb{F}$ and the synthetic one $\mathbb{F}^{(s)}$, which reads~\cite{josse2008testing}:

\begin{equation}
\label{eq:rv-def}
\mathrm{RV} \left( \mathbb{F}, \mathbb{F}^{(s)} \right) := \frac{\tr \left( S_{\mathbb{F} \mathbb{F}^{(s)}} S_{\mathbb{F}^{(s)} \mathbb{F}} \right)}{\sqrt{ \tr \left( S_{\mathbb{F} \mathbb{F}}^2 \right) \tr \left( S_{\mathbb{F}^{(s)} \mathbb{F}^{(s)}}^2 \right)}},
\end{equation}
using the following convention for any two matrices $\mathbb{A}$ and $\mathbb{B}$:
\begin{equation*}
S_{\mathbb{A} \mathbb{B}} = \frac{1}{n - 1} \mathbb{A}^\top \mathbb{B}.
\end{equation*}

The RV coefficient ranges from $0$ to $1$. It equals $1$ when the two data matrices are homothetic, that is when $\mathbb{F}^{(s)} = a \mathbb{F} + b$. Conversely, if all variables in the first matrix are uncorrelated from all variables in the second matrix, then $\mathrm{RV} = 0$. This metric is only applied to the \texttt{SynGait} approach because it relies on a one-to-one correspondence between individuals in the two data matrices which CTGAN and Copula synthesizers do not guarantee.

\subsubsection{Metrics from the Synthetic Data Vault}
\label{sec:sdv-metrics}

We selected two metrics from the SDV to evaluate the quality of the produced synthetic data. Each metric measures the similarity $\rho_k$ between individual columns of both data sets and the overall metric is defined as the average of the obtained similarities $\rho = \frac{1}{n-1} \sum_{k=1}^{n-1} \rho^{(k)}$.

\begin{description}
\item[\texttt{StatisticSimilarity}.] It measures the similarity between a real column and a synthetic column by comparing a summary statistic:
\begin{equation}
\label{eq:rho-fun}
\rho_\texttt{fun} := \frac{1}{n-1} \sum_{k=1}^{n-1} \max \left( 0, 1 - \frac{\left| \texttt{fun} \left( \mathbf{f}_k \right) - \texttt{fun} \left( \mathbf{f}_k^{(s)} \right) \right|}{\max \mathbf{f}_k - \min \mathbf{f}_k} \right),
\end{equation}
where $\mathrm{fun}$ is a statistical summary function. Specifically, we will use \texttt{mean} and \texttt{sd} therefore defining two metrics $\rho_\texttt{mean}$ and $\rho_\texttt{sd}$.

\item[\texttt{KSComplement}.] It measures the similarity between the marginal distributions of each column of the two data matrices as the complement of the Kolmogorov-Smirnov (KS) statistic. Specifically, the score reads:
\begin{equation}
\label{eq:rho-distr}
\rho_\texttt{distr} := \frac{1}{n - 1} \sum_{k=1}^{n-1} \left( 1 - \sup_{x \in \mathbb{R}} \left| F_k(x) - F_k^{(s)}(x) \right| \right),
\end{equation}
where $F_k$ and $F_k^{(s)}$ are the empirical distribution functions of the $k$-\textit{th} variable in both data sets.
\end{description}

All three metrics $\rho_\texttt{mean}$, $\rho_\texttt{sd}$ and $\rho_\texttt{distr}$ take values in $[0, 1]$ and a method performs well with respect to these metrics when they are close to one.

\subsubsection{Local cloakings and hidden rate}

Local cloakings and hidden rate aim at evaluating how good a method is at preserving privacy~\cite{avatar}. We are using these metrics to evaluate the risk of subject identification by mapping back synthetic data to the original data~\cite{b49}.

For each original observation, we compute the vector $\boldsymbol{\delta}_i \in \mathbb{R}^n$ of distances from the original functional scores to the synthetic scores of all observations:

\begin{equation*}
\delta_{ij} = \left\| \mathbf{f}_i - \mathbf{f}_j^{(s)} \right\| = \sqrt{\sum_{k=1}^{n-1} \left( f_{ik} - f_{jk}^{(s)} \right)^2}.
\end{equation*}

The \textbf{local cloaking} $\ell c_i$ of observation $i$ corresponds to the number of observations in the synthetic data set that are closer to $\mathbf{f}_i$ than $\mathbf{f}_i^{(s)}$. It reads:

\begin{equation*}
\ell c_i := \sum_{j=1}^n \left( \delta_{ij} < \delta_{ii} \right).
\end{equation*}

The higher this value, the better preserved the privacy of observation $i$ in the synthetic dataset. Local cloaking is however defined at the individual level. To get an overall sense of whether a method is good at preserving privacy, we can define the \textbf{hidden rate} $\mathrm{HR}$ which is the number of non-zero local cloakings and reads:

\begin{equation*}
\mathrm{HR} := \frac{1}{n} \sum_{i=1}^n \left( \ell c_i > 0 \right).
\end{equation*}
A method with high hidden rate is better at preserving privacy.

\section{Results}
\label{sec:results}

The Myo ancillary study is composed by 27 MS patient, their average EDSS score is 2. There are 10 patients with a score below 1.5 and 2 patients have a score over 5, the details of the EDSS distribution are presented in \Cref{tab:EDSS_MYO}. 

\begin{table}[!htbp]
\centering
\begin{tabular}{lcr}
EDSS & N & Prop. (\%) \\
\midrule[0.5pt]
0   & 7 & 25.93 \\
1   & 3 & 11.11 \\
1.5 & 1 &  3.70 \\
2   & 4 & 14.81 \\
2.5 & 5 & 18.52 \\
3   & 2 &  7.41 \\
4   & 3 & 11.11 \\
5.5 & 1 &  3.70 \\
6   & 1 &  3.70 \\
\end{tabular}
\caption{Distribution of EDSS scores for patients included in the MYO ancillary study}
\label{tab:EDSS_MYO}
\end{table}

\begin{figure}[!htpb]
\centering
\includegraphics[width=.7\linewidth]{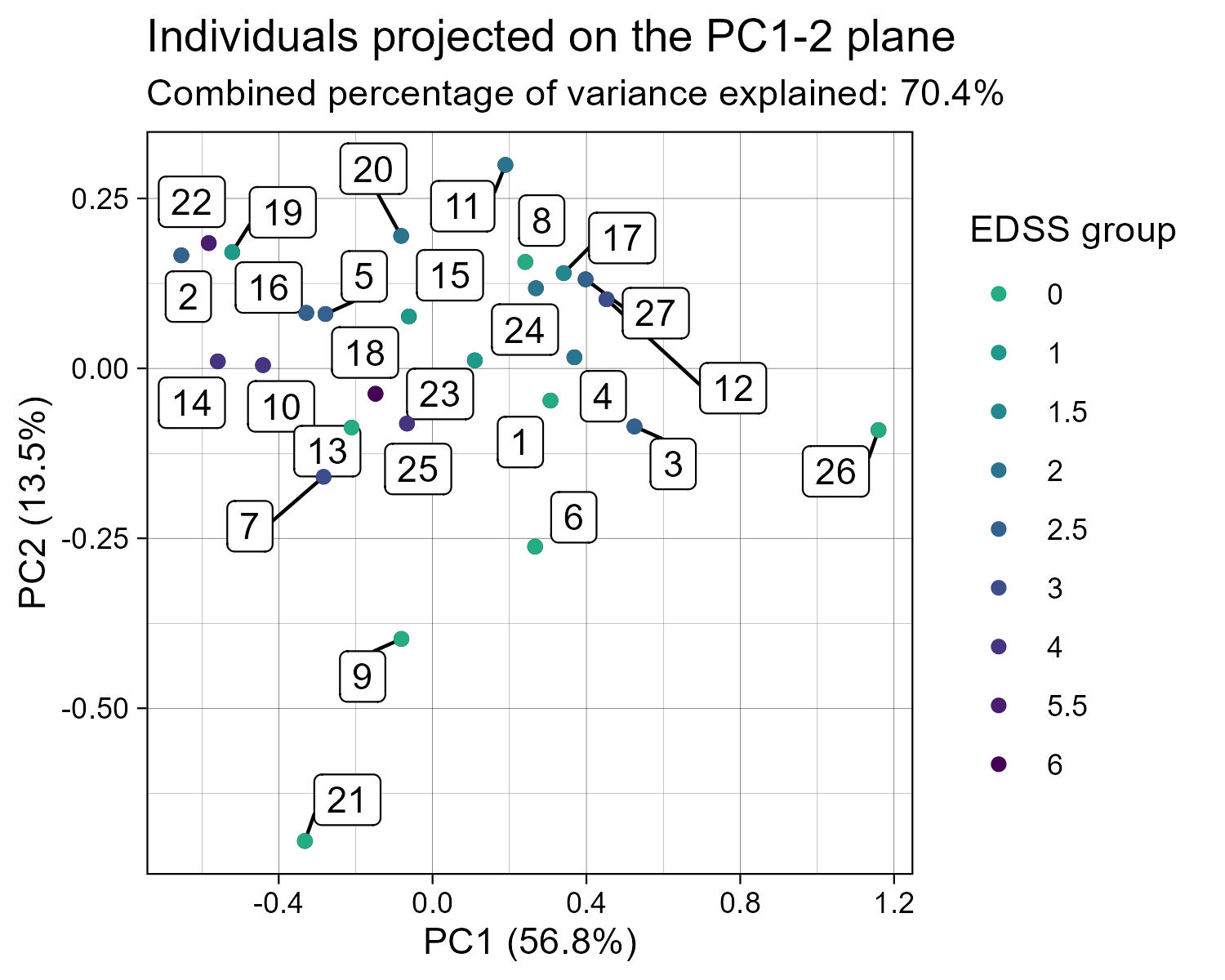}
\caption{Scatterplot of the first two principal components scores for the log-QFD of the $27$ individuals from the MYO study.}
\label{fig:FPCA}
\end{figure}

\Cref{fig:FPCA} displays the first two principal component's scores of the MFPCA applied on the centered log-qts of the MYO dataset, and colored by their EDSS. The first two dimensions of the MFPCA express $70.4\%$ of the log-QTS sample total inertia. Visualizing the scores helps understanding how the patients are characterized in regards to the first two mode of variation. It is also important to know the percentage of inertia of each principal mode of variation before selecting the parameters for synthetic data generation, and if there are some potential outliers. In \Cref{fig:FPCA}, one can observe some potential outliers (patients 21 and 26) that we can treat carefully in future analysis. 

In order to select the synthetic data generation hyper-parameters that would give us the best performance results in terms of fidelity we followed the process detailed in \Cref{sec:hpo}. The ranges of hyper-parameters values selected were:
\begin{itemize}
    \item $\alpha_0$: A sequence of 100 values between 0 and 50 with a logarithm growth. The cutoff was set at 50 because the Dirichlet distribution with higher values would very often be deterministic
    \item $\gamma$: A sequence of all integers between 2 and 8. This choice is explained by the size of the data set, weighting over 8 neighbors on a data set of size 27 does not make sense.
    \item $\tau$: A sequence of all possible number of components, here 1 to 26.
\end{itemize} 
For each combination of those hyper-parameters, 100 synthetic MFPCA scores were generated and distances $d_{\min}$ and $d_{\max}$ were computed (see \Cref{sec:hpo}). The threshold for the minimum distances was set to 0.024 which corresponds to 10 \% of the smallest distances from the original data set. In the appendix, we display the figures corresponding to the impact of the the parameters on $d_{\min}$ and $d_{\max}$. \Cref{f:pmin} shows the average minimum distances and \Cref{f:pmax} shows the average maximum distances. 

The selected parameters are finally $\alpha_0 = 4.52$, $\tau=9$, and $\gamma=2$. With those parameters, the mean maximum distance between synthetic scores $\Bar{d}_{\max}$ over the 100 data sets reaches 63\% of the maximum distance between two scores in the original data set, but if we omit the patient 26 who is a potential outlier, it reaches more than 90\%. The mean minimum distance  $\Bar{d}_{\min}$ is 10\% of the minimum distance between two scores in the original data. This is enough for us to consider that this synthetic data is truly different from the original one, and it guaranties us that two synthetic scores are not the same neither. 

\begin{figure}[!htpb]
\centering
\includegraphics[width=.99\linewidth]{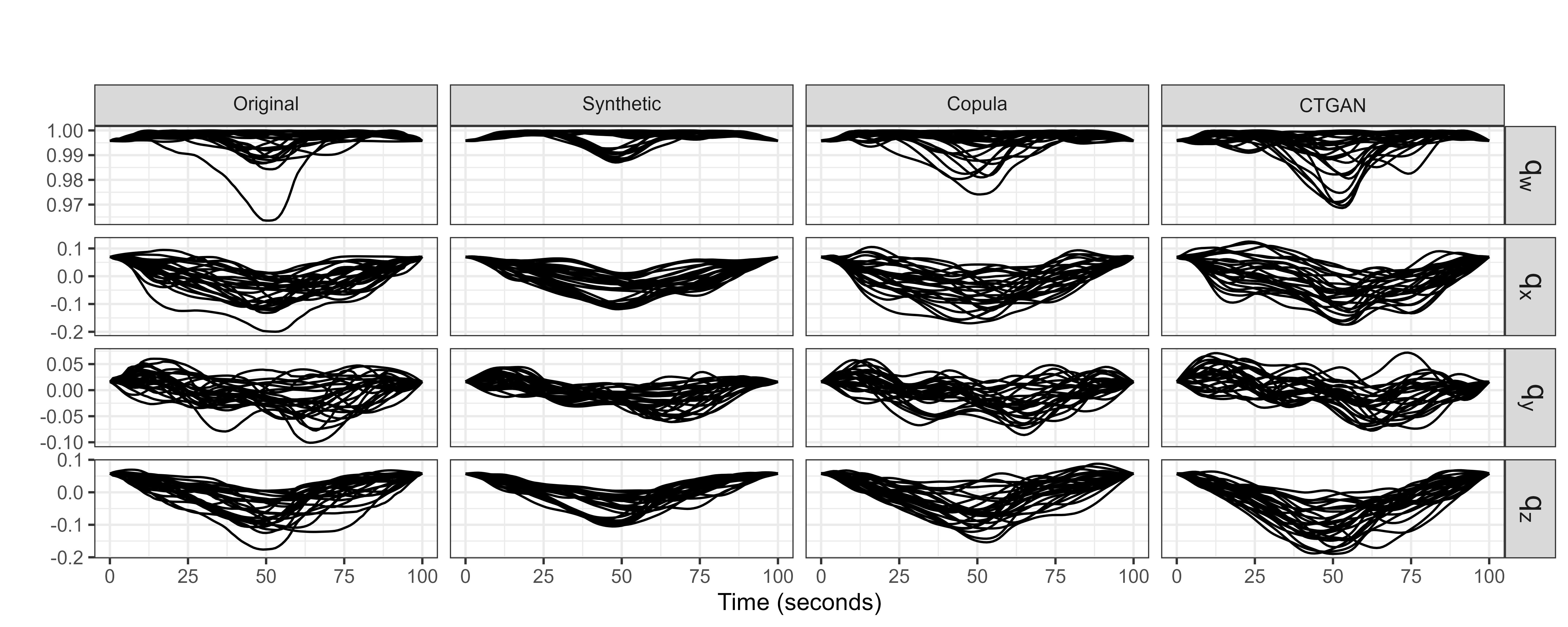}
\caption{\textbf{Generated individual gait patterns.} Using the original data (1st column), the proposed \texttt{SynGait} method (2nd column), the copula method (3rd column) and the CTGAN method (4th column).}
\label{fig:igps}
\end{figure}

\Cref{fig:igps} provides a representation of the original IGP data, and synthetic data obtained using the proposed framework with synthetic scores obtained by the \texttt{SynGait} method with the parameters previously chosen, the Copula synthesizer, and the CTGAN synthesizer. What strikes first looking at \Cref{fig:igps} is that the synthetic curves seems to be living on a smaller space. This is expected because the probability of having a strong weighting coefficient on an individual further from the others is small and the original dataset contains an outlier having a bigger amplitude than all the others. Except from that observation, the synthetic data is similar to the original data. 

Furthermore, while our data resides within the set of unit quaternions, we have yet to characterize the specific sub-variety where our IGPs truly lies. Employing methods such as Copula or GANs increases variability but introduces the risk of generating QTS that may not represent accurately multiple sclerosis patient's gait in real life. Besides, our method appears to reveal underlying groups in the original data more effectively, they are visible on the $\quat_x$,$\quat_y$, and $\quat_z$ components of the QTS.

\begin{figure}[!htpb]
\centering
\includegraphics[width=.99\linewidth]{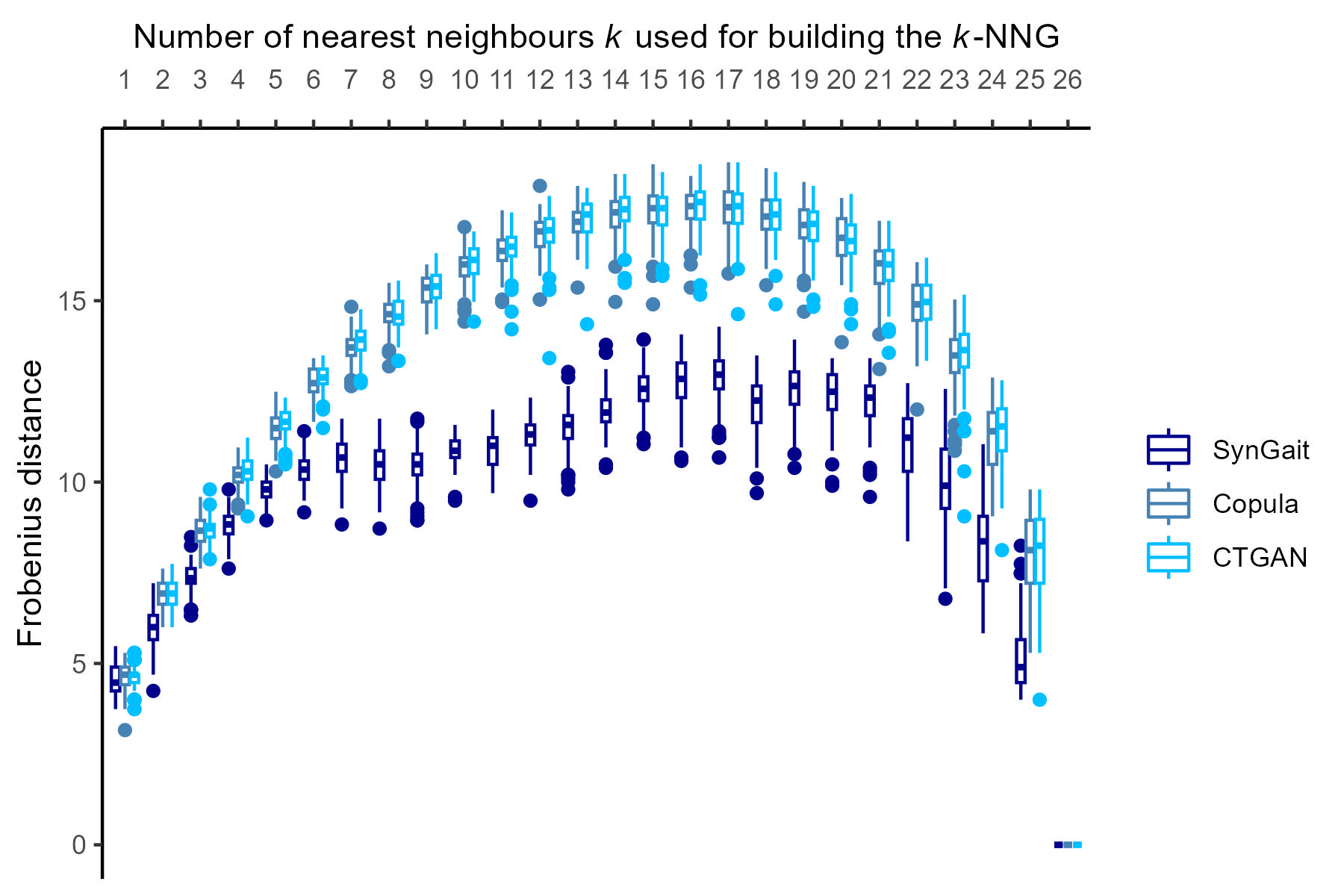}
\caption{\textbf{Distances between original manifold and synthesized ones.} Distribution of the Frobenius distances between the $k$-NNG built on the original IGPs and $k$-NNGs built from $100$ synthesized IGPs using the \texttt{SynGait} method (purple), the copula method (darkblue) and the CTGAN method (lightblue).}
\label{fig:knn}
\end{figure}

The primary quality that we wanted to verify was the superiority of our method in terms of geometric preservation. \Cref{fig:knn} illustrates the distribution of Frobenius distances (see \Cref{sec:fb}) between original $k$-NNGs and the synthetic ones for all possible values of $k$.

Our primary focus is the difference between the \texttt{SynGait} boxplots, which represent the Frobenius distances between the original score's graph and those generated by the proposed method. Those distances are compared with the Copula synthesizer and the CTGAN synthesizer integrated to the proposed framework to reduce gait data to tabular data.
We can easily notice than the distances are overall smaller with our method, excepted for low or high k values for the $k$-NNGs but those are not the values that allow to best capture the geometry of the data. This graph shows that our approach preserves well the geometry of the original data while creating synthetic IGPs.

In this setting, the mean RV coefficient is 0.84, indicating strong performance. This high RV coefficient demonstrates that our method effectively preserves the initial information present in the tabular dataset. 

\begin{figure}[!htpb]
\centering
\includegraphics[width=.85\linewidth]{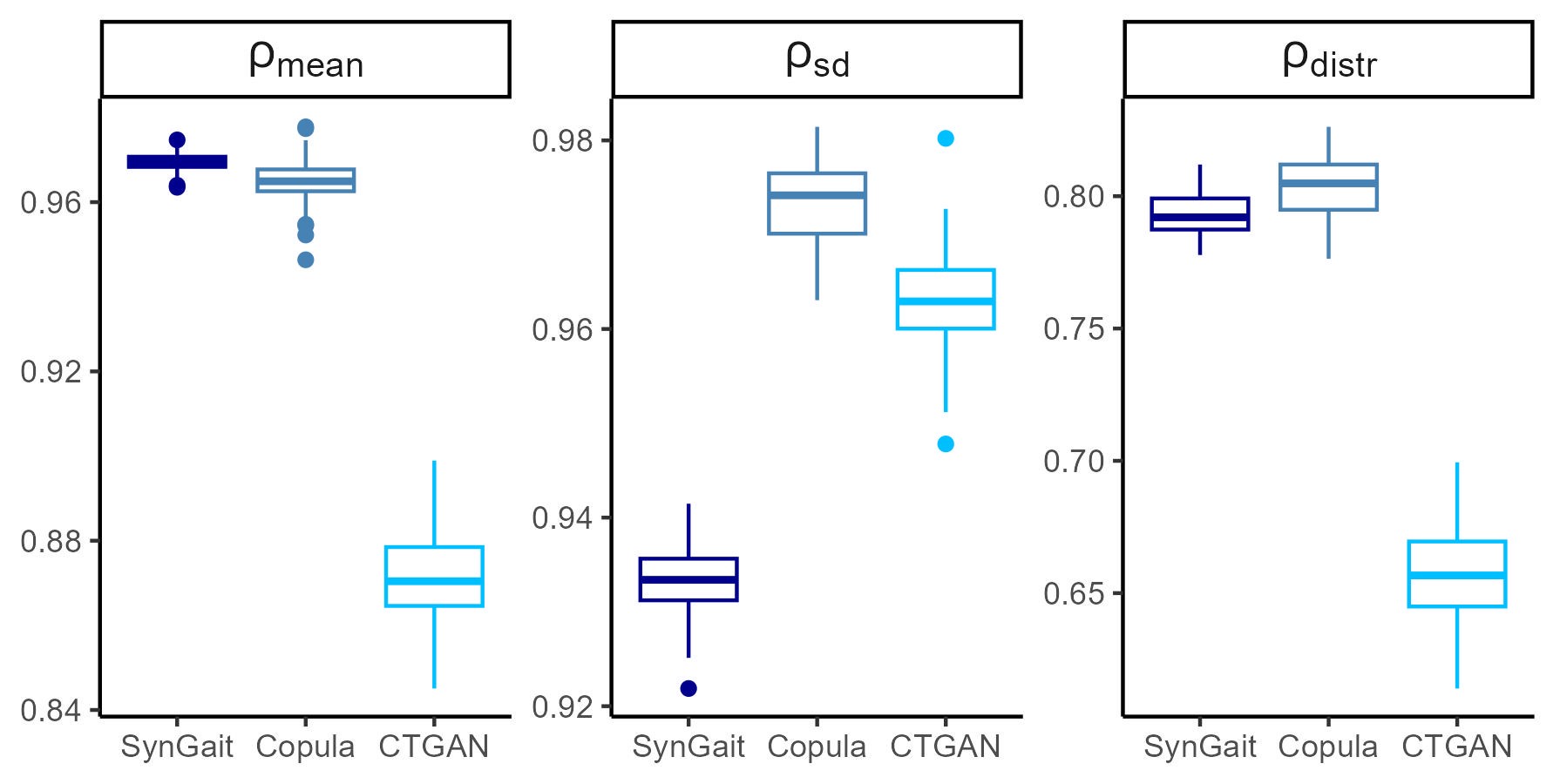}
\caption{\textbf{Distributions of performance metrics from the SDV.} Boxplots of the performance metrics $\rho_\mathrm{mean}$, $\rho_\mathrm{sd}$ and $\rho_\mathrm{distr}$ (\Cref{sec:sdv-metrics}) from the SDV colored by method. Higher values indicate better performance.}
\label{fig:metrics}
\end{figure}

\Cref{fig:metrics} illustrates the performances of the synthetic gait generation methods based on the metrics proposed by the Synthetic Data Vault. Notably, all three methods demonstrate strong performance results with most metrics results over 0.75. Overall the CTGAN is the one that performs worst, which could be explained by the fact that the original data set is quite small thus it's harder for machine learning based method to learn from the original data. The Copula synthesizer performs best on those fidelity metrics, followed  closely by \texttt{SynGait} which achieves high overall scores (and the best average $\rho_\mathrm{mean}$). However, \texttt{SynGait} is slightly less effective in matching the standard deviation with an average $\rho_\mathrm{sd}$ of 0.93. 

Finally, we are looking to avoid a small hidden rate to guarantee the anonymization of the synthetic data set. In the case of anonymization, only one of the hundred data sets generated is needed, we choose the one with the best local cloaking and hidden rate. With the previously chosen hyper-parameters, it corresponds to a local cloaking of 2.11 and a hidden rate of 85 \% which is satisfactory but we would advise to chose a higher number of neighbors to anonymize a data set.

\section{Discussion}
\label{s:discussion}

We have presented in this paper a novel method for the generation of synthetic unit quaternion time series characterizing human gait based on Multivariate Functional PCA (see \Cref{fig:FPCA}) on the tangent space of QTS and nearest-neighbors weighting. To the best of our knowledge, this is the first method to generate synthetic unit QTS. It provides interesting development perspectives for unit QTS studying and more particularly gait analysis using IGP. We additionally proposed an hyper-parameter optimization method focused on the coverage of the original inertia and the existence of a threshold ensuring that synthetic observations are truly new ones. 

We proposed evaluation tools to assess the quality of the produced synthetic data. This metrics are declined in two aspects, the fidelity and the privacy of the synthetic data. The first fidelity metric evaluates the geometry preservation, it is verified thanks to tools adapted from computational geometry. Other fidelity aspects are covered by metrics from the SDV and the RV coefficient. The \texttt{SynGait} method performed overall on all fidelity metrics and is particularly efficient to match the geometry of the data but lacks a little in matching the inertia of the original data, particularly if a single observation is far from all the other. Privacy is an important aspect of synthetic data as well, although we suppose that identifying a patient solely based on it's IGP is too challenging to be a real risk of identity disclosure, it has not been verified and this type of data can be considered as a quasi-identifier \cite{b29} thus one might want to make sure that privacy is not at risk before publishing synthetic gait data. The local cloaking and the hidden rate are privacy indicators representing the difficulty of mapping a synthetic observation back to the original one. In the synthetic data set generated the hidden rate of 85 \% is satisfactory, but as previously stated a higher number of neighbors would result in a better hidden rate. Generating a collection of synthetic data sets and concatenating them into a single bigger data sets also helps with anonymization.

We have shown a direct application to health studies with the generation of synthetic Individual Gait Patterns of multiple sclerosis patients. With this application, we proved that we are able to mimic the characteristics of a dataset while preserving the geometry of the unit QTS. This method can be applied to any human motion that can be described as unit QTS and to additional pathologies. Although one of the benefits of the \texttt{SynGait} method is that it does not need to train on a large volume of data, the advantage of the proposed framework is that any tabular data method can be implemented at the core of the algorithm to generate synthetic scores. With this framework, we have chosen to adapt two synthetic tabular data generation method to rotation data and evaluate them with the same metrics so we could use them as references. Those methods were chosen because of their popularity and open-source code \cite{SDV}. This implementation, using R for the MFPCA scores computation with the MFPCA package \cite{b14} and for the \texttt{SynGait} algorithm and Pyhton for the SDV tools, allowed a comparison between our approach, a GAN approach and a Copula approach, highlighting that the true strenght of the \texttt{SynGait} method lies in the geometry conservation, and a very good fidelity as well.

The drawback of the \texttt{SynGait} method is the running time of the optimization algorithm. The setting we presented in \Cref{sec:results} implies the generation of 1820000 synthetic scores and computation of associated distances. It takes about 8 hours and 50 minutes to run on a DELL computer running under windows with i7-1185G7 processor with 7 threads at a 3.0 GHz frequency. From a user's perspective, it is not needed to be as precise as we were and one can choose less parameters to test, and less repetitions as the results are already quite stable after 10 to 30 repetitions. We also worked on default setting that we validated on two other unit QTS data sets of bigger sizes (respectively 39 and 64 observations) and we suggest setting a number of components from the MFPCA that represents around 95 \% of the variability of the data, $\alpha_0$ set to 5 and a number of neighbors close $1/10^{th}$ of the number of observations.

This method will allow us to pursue our work around the IGP and the understanding of those gait patterns for multiple sclerosis patients. After extending this work to mix data, we will be able to generate synthetic EDSS scores associated to the synthetic QTS and use those synthetic data to test the robustness of clustering algorithms. This method generates a synthetic data set of the same dimension as the original one, but we will use a collection of synthetic data sets to use as a form of bootstrapping for clustering results. Clustering on the MYO dataset has already been published by Drouin et al. \cite{b25} and promising results were shared. When the stability of the clustering will be proven, we will use those interpretable groups as reference labels then affect future data to them.

\section{Acknowledgements}
The authors would like to thank the France Sclérose en Plaques (FSP) foundation (formerly ARSEP) and the Agence pour les Mathématiques en Interaction avec l'Entreprise et la Société (AMIES) for funding the clinical studies, as well as the Observatoire Français de la Sclérose en Plaques (OFSEP), the teaching university hospital of Nantes and Prof. P.A. Gourraud for facilitating the ancillary study on the MYO study. This work is part of a Ph.D. thesis co-financed by the ANR AIBY4 (ANR-20-THIA-0011) and Nantes University.
\appendix
\newpage
\section{Appendix}
\label{s:ap}
\begin{figure*}[!htbp]
\centerline{\includegraphics[width=7.5 in]{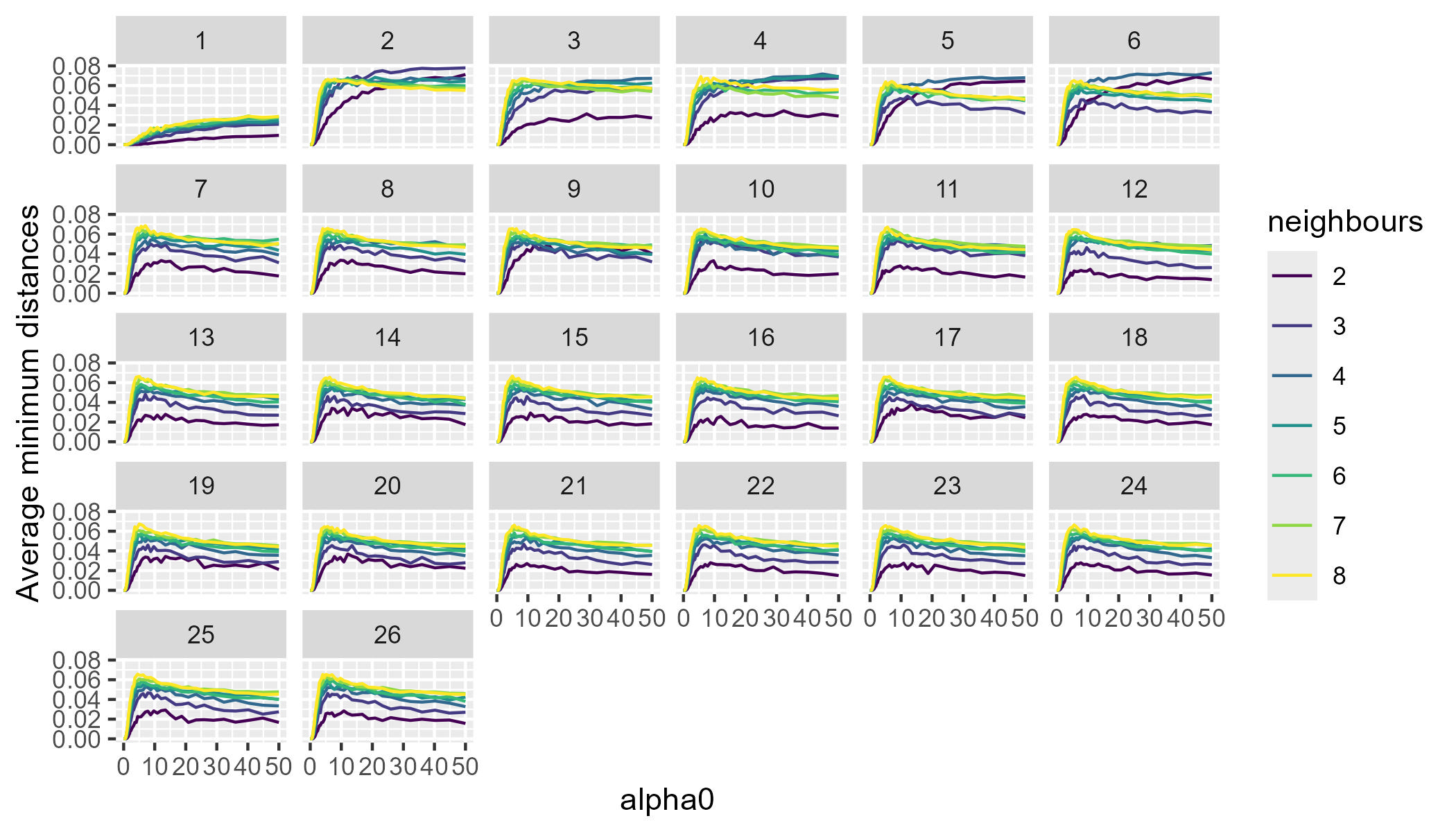}}
\caption{Graph of the minimum distances between synthetic and original individuals}
\label{f:pmin}
\end{figure*}

\begin{figure*}[!htbp]
\centerline{\includegraphics[width=7.5 in]{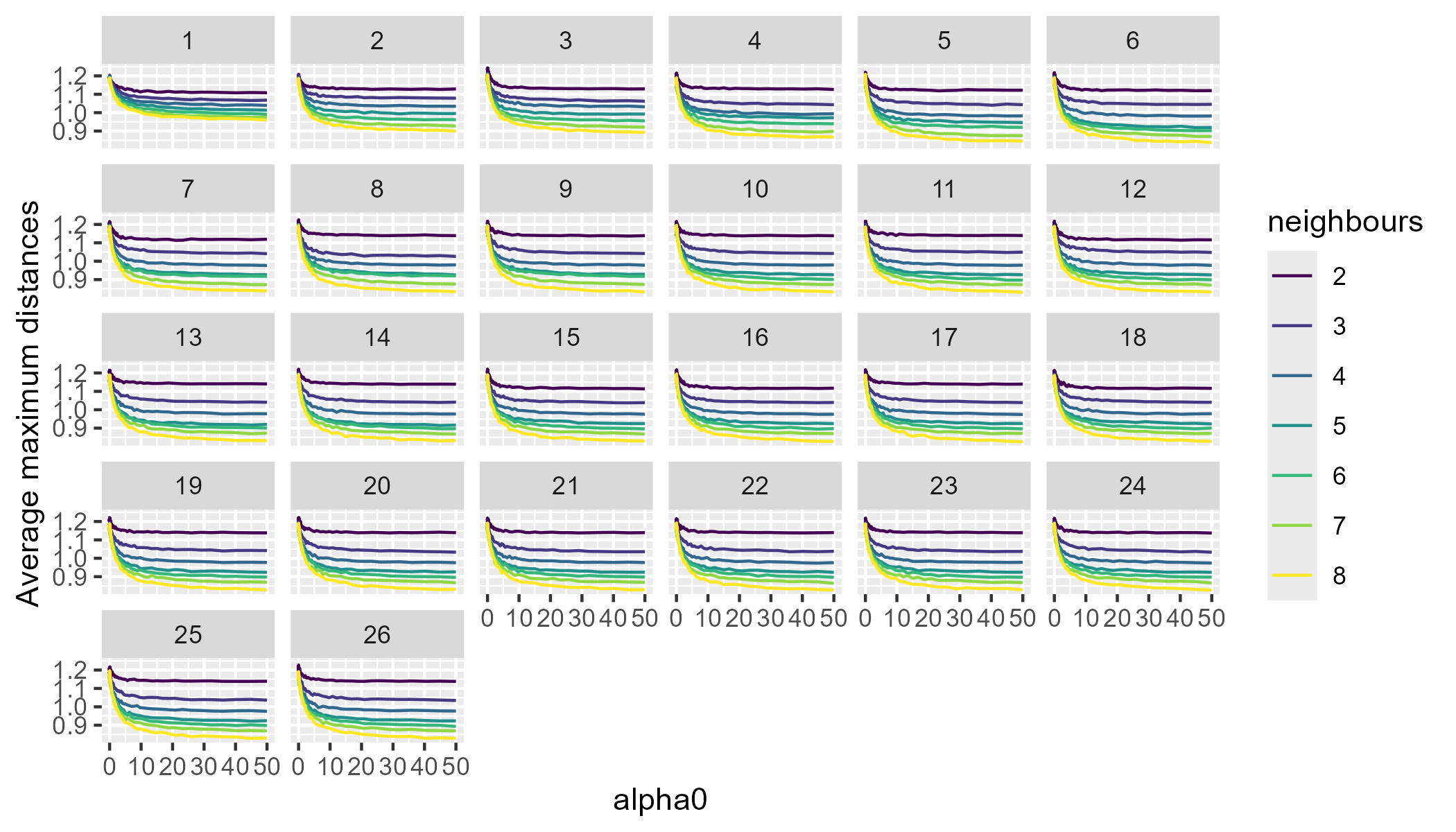}}
\caption{Graph of the maximum distances between synthetic individuals}
\label{f:pmax}
\end{figure*}

\newpage


\bibliographystyle{unsrt}
\bibliography{biblio.bib}

\begin{thebibliography}{10}

\bibitem{MS_symptoms}
Ilya Kister, Tamar~E. Bacon, Eric Chamot, Amber~R. Salter, Gary~R. Cutter, Jennifer~T. Kalina, and Joseph Herbert.
\newblock {Natural History of Multiple Sclerosis Symptoms}.
\newblock {\em International Journal of MS Care}, 15(3):146--156, 10 2013.

\bibitem{LaRocca_2011}
Nicholas~G. LaRocca.
\newblock Impact of walking impairment in multiple sclerosis.
\newblock {\em The Patient: Patient-Centered Outcomes Research}, 4(3):189–201, September 2011.

\bibitem{b42}
L~Rocher, J.~M. Hendrickx, and Y~de~Montjoye.
\newblock Estimating the success of re-identifications in incomplete datasets using generative models.
\newblock {\em Nature Communications}, 10, 2019.

\bibitem{b43}
Mark~A. Rothstein.
\newblock Is deidentification sufficient to protect health privacy in research?
\newblock {\em The American Journal of Bioethics}, 10(9):3--11, 2010.

\bibitem{b44}
Arvind Narayanan and Edward~W Felten.
\newblock No silver bullet: De-identification still doesn’t work.
\newblock {\em White Paper}, 8, 2014.

\bibitem{Wang24}
Zhenchen Wang, Barbara Draghi, Ylenia Rotalinti, Darren Lunn, and Puja Myles.
\newblock High-fidelity synthetic data applications for data augmentation.
\newblock In Manuel Domínguez-Morales, Javier Civit-Masot, Luis Muñoz-Saavedra, and Robertas Damaševičius, editors, {\em Deep Learning}, chapter~7. IntechOpen, Rijeka, 2024.

\bibitem{b11}
Donald~B. Rubin.
\newblock Statistical disclosure limitation: discussion.
\newblock {\em Journal of Official Statistics}, 9:2461--468, 1993.

\bibitem{raghunathan2003}
Trivellore~E Raghunathan, Jerome~P Reiter, and Donald~B Rubin.
\newblock Multiple imputation for statistical disclosure limitation.
\newblock {\em Journal of official statistics}, 19(1):1, 2003.

\bibitem{VersatileGait}
Pengyi Zhang, Huanzhang Dou, Wenhu Zhang, Yuhan Zhao, Zequn Qin, Dongping Hu, Yi~Fang, and Xi~Li.
\newblock A large-scale synthetic gait dataset towards in-the-wild simulation and comparison study.
\newblock {\em ACM Trans. Multimedia Comput. Commun. Appl.}, 19(1), 1 2023.

\bibitem{ParkisonGait}
Jorge Marquez~Chavez and Wei Tang.
\newblock A vision-based system for stage classification of parkinsonian gait using machine learning and synthetic data.
\newblock {\em Sensors}, 22(12), 2022.

\bibitem{Kim_Hargrove_2023}
Minjae Kim and Levi~J. Hargrove.
\newblock Generating synthetic gait patterns based on benchmark datasets for controlling prosthetic legs.
\newblock {\em Journal of NeuroEngineering and Rehabilitation}, 20(1), 9 2023.

\bibitem{ICF}
World~Health Organization.
\newblock International classification of functioning, disability, and health: Icf\_2001, 2001.

\bibitem{PressureMat2017}
Vidya~K. Nandikolla, Robin Bochen, Steven Meza, and Allan Garcia.
\newblock Experimental gait analysis to study stress distribution of the human foot.
\newblock {\em Journal of Medical Engineering}, 2017(1):3432074, 2017.

\bibitem{SilhouetteGait2006}
Ju~Han and Bir Bhanu.
\newblock Individual recognition using gait energy image.
\newblock {\em IEEE Trans Pattern Anal Mach Intell}, 28(2):316--22, 2006.

\bibitem{Vicon2017}
H~Lamine, S~Bennour, M~Laribi, L~Romdhane, and S~Zaghloul.
\newblock Evaluation of calibrated kinect gait kinematics using a vicon motion capture system.
\newblock {\em Computer Methods in Biomechanics and Biomedical Engineering}, 20(sup1):S111--S112, 2017.

\bibitem{sensorWeijun2012}
Weijun Tao, Tao Liu, Rencheng Zheng, and Hutian Feng.
\newblock Gait analysis using wearable sensors.
\newblock {\em Sensors}, 12(2):2255--2283, 2012.

\bibitem{T25FW2012}
Bernd~C. Kieseier and Carlo Pozzilli.
\newblock Assessing walking disability in multiple sclerosis.
\newblock {\em Multiple Sclerosis Journal}, 18(7):914–924, 2012.

\bibitem{b34}
John~F. Kurtzke.
\newblock Rating neurologic impairment in multiple sclerosis: an expanded disability status scale (edss).
\newblock {\em Neurology}, 33(11), 1983.

\bibitem{b30}
Pierre Drouin, Aymeric Stamm, Laurent Chevreuil, Vincent Graillot, Laetitia Barbin, Philippe Nicolas, et~al.
\newblock Gait impairment monitoring in multiple sclerosis using a wearable motion sensor.
\newblock {\em Medical Case reports and Reviews}, 5:1--5, 2022.

\bibitem{b16}
John Voight.
\newblock {\em Quaternion Algebras}.
\newblock Springer Nature, 2005.

\bibitem{b32}
Mathijs~S. Dijkhuizen.
\newblock The double covering of the quantum group soq(3).
\newblock In {\em Proceedings of the Winter School "Geometry and Physics". Circolo Matematico di Palermo}, volume~37, pages 47--57, 1994.

\bibitem{b9}
James~O. Ramsay and Bernard~W. Silverman.
\newblock {\em Principal components analysis for functional data}, pages 147--172.
\newblock Springer New York, 2005.

\bibitem{avatar}
Morgan Guillaudeux, Olivia Rousseau, Julien Petot, Zineb Bennis, Charles-Axel Dein, Thomas Goronflot, et~al.
\newblock Patient-centric synthetic data generation, no reason to risk re-identification in biomedical data analysis.
\newblock {\em npj Digital Medicine}, 6, 2023.

\bibitem{sola2018micro}
Joan Sola, Jeremie Deray, and Dinesh Atchuthan.
\newblock A micro lie theory for state estimation in robotics.
\newblock {\em arXiv preprint arXiv:1812.01537}, 2018.

\bibitem{b40}
James~O. Ramsay and Bernard~W. Silverman.
\newblock {\em Introduction}, pages 1--18.
\newblock Springer New York, 2005.

\bibitem{b7}
K.~Pearson.
\newblock Liii. on lines and planes of closest fit to systems of points in space.
\newblock {\em The London, Edinburgh, and Dublin Philosophical Magazine and Journal of Science}, 2(11):559--572, 1901.

\bibitem{b8}
Harold Hotelling.
\newblock Analysis of a complex of statistical variables into principal components.
\newblock {\em Journal of Educational Psychology}, 24:498--520, 1933.

\bibitem{b38}
Clara Happ and Sonja Greven.
\newblock Multivariate functional principal component analysis for data observed on different (dimensional) domains.
\newblock {\em Journal of the American Statistical Association}, 113(522):649--659, 2018.

\bibitem{b14}
Clara Happ-Kurz.
\newblock {\em MFPCA: Multivariate Functional Principal Component Analysis for Data Observed on Different Dimensional Domains}, 2022.
\newblock R package version 1.3-10.

\bibitem{b37}
{Kai Wang} Ng, {Guo Liang} Tian, and {Man Lai} TANG.
\newblock {\em Dirichlet and Related Distributions: Theory, Methods and Applications}.
\newblock Wiley-Blackwell, 2011.

\bibitem{SDV}
Neha Patki, Roy Wedge, and Kalyan Veeramachaneni.
\newblock The synthetic data vault.
\newblock In {\em 2016 IEEE International Conference on Data Science and Advanced Analytics (DSAA)}, pages 399--410, 2016.

\bibitem{b56}
Kevin Zhang, Neha Patki, and Kalyan Veeramachaneni.
\newblock Sequential models in the synthetic data vault.
\newblock {\em arXiv preprint arXiv:2207.14406}, 2022.

\bibitem{b61}
Shih-Chieh Kao, Hoe~Kyoung Kim, Cheng Liu, Xiaohui Cui, and Budhendra~L. Bhaduri.
\newblock Dependence-preserving approach to synthesizing household characteristics.
\newblock {\em Transportation Research Record}, 2302(1):192--200, 2012.

\bibitem{b62}
Yi~Sun, Alfredo Cuesta-Infante, and Kalyan Veeramachaneni.
\newblock Learning vine copula models for synthetic data generation.
\newblock {\em Proceedings of the AAAI Conference on Artificial Intelligence}, 33(01):5049--5057, 7 2019.

\bibitem{sklar1959fonctions}
M~Sklar.
\newblock Fonctions de r{\'e}partition {\`a} n dimensions et leurs marges.
\newblock In {\em Annales de l'ISUP}, volume~8, pages 229--231, 1959.

\bibitem{meyer2021copula}
David Meyer, Thomas Nagler, and Robin~J Hogan.
\newblock Copula-based synthetic data generation for machine learning emulators in weather and climate: application to a simple radiation model.
\newblock {\em Geoscientific Model Development Discussions}, 2021:1--21, 2021.

\bibitem{b52}
Ian Goodfellow, Jean Pouget-Abadie, Mehdi Mirza, Bing Xu, David Warde-Farley, Sherjil Ozair, Aaron Courville, and Yoshua Bengio.
\newblock Generative adversarial networks.
\newblock {\em Communications of the ACM}, 63(11):139--144, 2020.

\bibitem{GAN_overview2018}
Antonia Creswell, Tom White, Vincent Dumoulin, Kai Arulkumaran, Biswa Sengupta, and Anil~A. Bharath.
\newblock Generative adversarial networks: An overview.
\newblock {\em IEEE Signal Processing Magazine}, 35(1):53--65, 2018.

\bibitem{xu2019modeling}
Lei Xu, Maria Skoularidou, Alfredo Cuesta-Infante, and Kalyan Veeramachaneni.
\newblock Modeling tabular data using conditional gan.
\newblock {\em Advances in neural information processing systems}, 32, 2019.

\bibitem{ComputationalGeom2012}
Franco~P Preparata and Michael~I Shamos.
\newblock {\em Computational geometry: an introduction}.
\newblock Springer Science \& Business Media, 2012.

\bibitem{b23}
David~J. Marchette.
\newblock {\em cccd: Class Cover Catch Digraphs}, 2022.
\newblock R package version 1.6.

\bibitem{josse2008testing}
J.~Josse, J.~Pagès, and F.~Husson.
\newblock Testing the significance of the rv coefficient.
\newblock {\em Computational Statistics \& Data Analysis}, 53(1):82--91, 2008.

\bibitem{b35}
P.~Robert and Y.~Escoufier.
\newblock A unifying tool for linear multivariate statistical methods: The rv- coefficient.
\newblock {\em Journal of the Royal Statistical Society. Series C (Applied Statistics)}, 25(3):257--265, 1976.

\bibitem{b49}
Stacey Truex, Ling Liu, Mehmet~Emre Gursoy, Lei Yu, and Wenqi Wei.
\newblock Demystifying membership inference attacks in machine learning as a service.
\newblock {\em IEEE Transactions on Services Computing}, 14:2073--2089, 2019.

\bibitem{b29}
Abdul Majeed and Sungchang Lee.
\newblock Anonymization techniques for privacy preserving data publishing: A comprehensive survey.
\newblock {\em IEEE Access}, 9:8512--8545, 2021.

\bibitem{b25}
Pierre Drouin, Aymeric Stamm, Laurent Chevreuil, Vincent Graillot, Laetitia Barbin, Pierre‐Antoine Gourraud, et~al.
\newblock Semi‐supervised clustering of quaternion time series: Application to gait analysis in multiple sclerosis using motion sensor data.
\newblock {\em Statistics in Medicine}, 42(4):433–456, 2022.

\end{thebibliography}







\end{document}